\documentclass[10pt,journal,compsoc]{IEEEtran}
\ifCLASSOPTIONcompsoc
  \usepackage[nocompress]{cite}
\else
  \usepackage{cite}
\fi

\usepackage[utf8]{inputenc}
\usepackage{amsfonts, amssymb, amsmath, amsthm}
\usepackage{dsfont}
\usepackage{graphicx}
\usepackage{multirow}
\usepackage{array}
\usepackage{hyperref}
\usepackage{wrapfig}
\usepackage{booktabs}

\usepackage{algorithm}
\usepackage{algpseudocode}

\usepackage{multicol}

\usepackage{xcolor}
\usepackage{colortbl}

\usepackage{pgf,tikz,pgfplots,pgfplotstable}
\usetikzlibrary{plotmarks}
\usetikzlibrary{shapes,arrows}
\usepackage{pgfmath}
\usepgfplotslibrary{statistics}
\usepgfplotslibrary{groupplots}

\usepackage{balance}
\usepackage{eucal}

\newcommand{\arl}[1]{\text{ARL}_{#1}}

\newtheorem{proposition}{Proposition}
\pgfplotsset{compat=1.15}
\begin{document}
\definecolor{color_qt}{rgb}{.5,.8,.3}
\definecolor{color_l}{rgb}{.2,.6,.8}
\definecolor{color_h}{rgb}{1,.6,.1}

\definecolor{our_blue}{rgb}{.2,.6,.8}
\definecolor{our_orange}{rgb}{1,.5,.05}
\definecolor{our_green}{rgb}{.17,.63,.17}
\definecolor{our_red}{rgb}{.84,.15,.16}
\definecolor{our_purple}{rgb}{.58,.4,.74}
\definecolor{our_msblue}{rgb}{.48,.41,.93}
\definecolor{our_mblue}{rgb}{0,0,1}
\definecolor{our_darkblue}{rgb}{0,0,.55}
\definecolor{our_indigo}{rgb}{.29,0,.51}

\pgfplotsset{
    style_stop/.style={
        width=0.24\textwidth,
        height=0.26\columnwidth,
        scale only axis,
        xtick = {250,500,750,1000}, 
        title style = {
            font = \scriptsize,
            at = {(0.5, 0.95)},
        },
        xlabel style = {
        	align=center,
        	anchor = near ticklabel, 
        	at = {(0.5,-0.1)},
            font=\scriptsize,
        },
        xticklabel style = {font = \tiny},
        yticklabel style = {font = \tiny},
        ylabel style = {
            anchor = center, 
            at = {(-0.18,0.5)},
            font=\scriptsize,
        },
        /pgf/number format/.cd,
        1000 sep={},
    }
}

\pgfplotsset{
    style_auc/.style={
        width=0.24\textwidth,
        height=0.26\columnwidth,
        scale only axis,
        xtick = {1,2,3,4,5},
        xticklabels = {$50$,$100$,$200$,$500$,$1000$}, 
        title style = {
            font = \scriptsize,
            at = {(0.5, 0.95)},
        },
        xlabel style = {
        	align=center,
        	anchor = near ticklabel, 
        	at = {(0.5,-0.1)},
            font=\scriptsize,
        },
        xticklabel style = {font = \tiny},
        yticklabel style = {font = \tiny},
        ylabel style = {
            anchor = center, 
            at = {(-0.18,0.5)},
            font=\scriptsize,
        },
        ymajorgrids=true,
    }
}

\pgfplotsset{
    style_skl/.style={
        width=0.24\textwidth,
        height=0.13\textwidth,
        scale only axis,
        xtick = {0.5,1,...,3}, 
        xticklabels = {$0.5$,$1$,$1.5$,$2$,$2.5$,$3$},
        title style = {
            font = \scriptsize,
            at = {(0.5, 0.95)},
        },
        xlabel style = {
        	align=center,
        	anchor = near ticklabel, 
        	at = {(0.5,-0.1)},
            font=\scriptsize,
        },
        xticklabel style = {font = \tiny},
        yticklabel style = {font = \tiny},
        ylabel style = {
            anchor = center, 
            at = {(-0.18,0.5)},
            font=\scriptsize,
        },
    }
}

\pgfplotsset{
    style_bins/.style={
        width=0.38\textwidth,
        height=0.33\columnwidth,
        scale only axis,
        xtick = {0,1,...,8},
        xticklabels = {$2$,$4$,$8$,$16$,$32$,$64$,$128$,$256$,$512$},
        title style = {
            font = \scriptsize,
            at = {(0.5, 0.95)},
        },
        xlabel style = {
        	align=center,
        	anchor = near ticklabel, 
        	at = {(0.5,-0.1)},
            font=\scriptsize,
        },
        xticklabel style = {font = \tiny},
        yticklabel style = {font = \tiny},
        ylabel style = {
            anchor = center, 
            at = {(-0.12,0.5)},
            font=\scriptsize,
        },
    }
}

\pgfplotsset{
    style_arl/.style={
        width=0.19\textwidth,
        height=0.135\textwidth,
        scale only axis,
        scaled ticks=false,
        xmin = 175, 
        xmax = 5375,
        xtick = {500,1000,2000,5000},
        xticklabels = {$500\;\;$,$\;\;1000$,$\;2000$,$5000$},
        title style = {
            font = \scriptsize,
            at = {(0.5, 0.95)},
        },
        xlabel style = {
        	align=center,
        	anchor = near ticklabel, 
        	at = {(0.5,-0.1)},
            font=\scriptsize,
        },
        xticklabel style = {font = \tiny},
        yticklabel style = {font = \tiny},
        ylabel style = {
            anchor = center, 
            at = {(-0.22,0.5)},
            font=\scriptsize,
        },
        /pgf/number format/.cd,
        use comma,
        1000 sep={},
    }
}

\pgfplotsset{
    style_delay/.style={
        width=0.19\textwidth,
        height=0.135\textwidth,
        scale only axis,
        xtick = {0,10,...,80}, 
        minor xtick = {63, 39, 22, 9.5},
        title style = {
            font = \scriptsize,
            at = {(0.5, 0.95)},
        },
        xlabel style = {
        	align=center,
        	anchor = near ticklabel, 
        	at = {(0.5,-0.1)},
            font=\scriptsize,
        },
        xticklabel style = {font = \tiny},
        yticklabel style = {font = \tiny},
        ylabel style = {
            anchor = center, 
            at = {(-0.22,0.5)},
            font=\scriptsize,
        },
        /pgf/number format/.cd,
        use comma,
        1000 sep={},
    }
}

\pgfplotsset{
    style_fig_del_sim/.style={
        width=0.17\textwidth,
        height=0.23\columnwidth,
        scale only axis,
        xmin = 0.2, 
        xmax = 3.3,
        xtick = {0.5,1,...,3}, 
        title style = {
            font = \scriptsize,
            at = {(0.5, 0.9)},
        },
        xlabel style = {
        	align=center,
        	anchor = near ticklabel, 
        	at = {(0.5,-0.15)},
            font=\footnotesize,
        },
        xticklabel style = {font = \scriptsize},
        yticklabel style = {font = \scriptsize},
        ylabel style = {
            anchor = center, 
            at = {(-0.25,0.5)},
            font=\footnotesize,
        },
    }
}

\pgfplotsset{
    style_fig_del_real/.style={
        width=0.175\textwidth,
        height=0.31\columnwidth,
        scale only axis,
        xmin = 0, 
        xmax = 1,
        xtick = {0.001,0.01,0.1,1}, 
        title style = {
            font = \small,
            at = {(0.5, 0.9)},
        },
        xlabel style = {
        	align=center,
        	anchor = near ticklabel, 
        	at = {(0.5,-0.1)},
            font=\footnotesize,
        },
        xticklabel style = {font = \scriptsize},
        yticklabel style = {font = \scriptsize},
        ylabel style = {
            anchor = center, 
            at = {(-0.25,0.5)},
            font=\footnotesize,
        },
    }
}

\pgfplotsset{
    style_line/.style={
        line width = 1pt,
    },
}

\pgfplotsset{
    style_qtewma/.style={
        style_line,
        color = our_blue,
        mark = square,
    },
    style_qt/.style={
        style_line,
        color = our_orange,
        mark = o,
    },
    style_spll/.style={
        style_line,
        color = our_red,
        mark = asterisk,
    },
    style_cpm/.style={
        style_line,
        color = our_green,
        mark = diamond,
    },
    style_scanb/.style={
        style_line,
        color = our_purple,
        mark = triangle,
    },
    style_truep/.style={
        dashed,
        style_line,
        color = our_orange,
        mark = star,
    },
    style_beta2/.style={
        style_line,
        color = our_green,
        mark = halfsquare*,
    },
    style_beta5/.style={
        style_line,
        color = our_mblue,
        mark = pentagon,
    },
    style_beta10/.style={
        style_line,
        color = our_red,
        mark = halfcircle*,
    },
    style_stop512/.style={
        style_line,
        color = our_green,
        mark = x,
    },
    style_stop1024/.style={
        style_line,
        color = our_red,
        mark = triangle*,
    }
}


\title{Nonparametric and Online Change Detection in Multivariate Datastreams using QuantTree}
\author{Luca Frittoli, Diego Carrera, Giacomo Boracchi}

\markboth{Accepted for publication in IEEE Transactions on Knowledge and Data Engineering. \copyright 2022 IEEE DOI: 10.1109/TKDE.2022.3201635}%
{Frittoli \MakeLowercase{\textit{et al.}}: QT-EWMA-update}


\IEEEtitleabstractindextext{%
\begin{abstract}
We address the problem of online change detection in multivariate datastreams, and we introduce QuantTree Exponentially Weighted Moving Average (QT-EWMA), a nonparametric change-detection algorithm that can control the expected time before a false alarm, yielding a desired Average Run Length (ARL$_0$). Controlling false alarms is crucial in many applications and is rarely guaranteed by online change-detection algorithms that can monitor multivariate datastreams without knowing the data distribution. Like many change-detection algorithms, QT-EWMA builds a model of the data distribution, in our case a QuantTree histogram, from a stationary training set. To monitor datastreams even when the training set is extremely small, we propose QT-EWMA-update, which incrementally updates the QuantTree histogram during monitoring, always keeping the ARL$_0$ under control. Our experiments, performed on synthetic and real-world datastreams, demonstrate that QT-EWMA and QT-EWMA-update control the ARL$_0$ and the false alarm rate better than state-of-the-art methods operating in similar conditions, achieving lower or comparable detection delays.
\end{abstract}

\begin{IEEEkeywords}
online change detection, nonparametric monitoring, multivariate datastreams, histograms, false alarms.
\end{IEEEkeywords}}

\maketitle

\IEEEdisplaynontitleabstractindextext
\IEEEpeerreviewmaketitle


\IEEEraisesectionheading{
\section{Introduction}\label{sec:introduction}
}
\IEEEPARstart{C}{hange} detection in datastreams \cite{aggarwal2005change} is a challenging problem with relevant applications in many domains including quality control \cite{hawkins2003changepoint}, security \cite{tartakovsky2006novel}, cryptographic attacks \cite{frittoli2020strengthening}, finance \cite{Ross2011}, and in several engineering problems, where control charts have been employed for decades~\cite{basseville1993}. A relevant example is industrial process monitoring, where production machinery is equipped with multiple sensors that measure vibration frequency, flow, temperature, pressure, etc. These observations form datastreams that must be monitored since any distribution change might indicate failures or ongoing deterioration of specific components such as bearings and gears, thus change detection can be key for predictive maintenance~\cite{popescu2021basic}. Change detection is also studied in the machine learning literature since classification and recommendation systems often operate on streaming data. Here, changes are called \emph{concept drifts} and classifiers must be adapted to an evolving data-generating process \cite{lu2018learning}.

Many of these applications require \emph{multivariate} datastreams to be processed \emph{online}, i.e., while acquiring new observations. This condition represents a crucial challenge when designing and implementing change-detection algorithms. On the one hand, a device implementing a change-detection algorithm has limited memory and can perform a limited number of operations at each time $t$, while datastreams are virtually unlimited. On the other hand, to increase the detection power, online change-detection algorithms would need in principle to analyze, at each time $t$, all the data observed until $t$, and this typically implies an increase of computational and memory requirements~\cite{Ross2011}. Another fundamental challenge is to monitor multivariate datastreams in a \emph{nonparametric} manner, which enables change-detection algorithms to operate when the initial distribution $\phi_0$ is unknown. Unfortunately, most nonparametric online change-detection algorithms can only monitor \emph{univariate} datastreams \cite{Ross2011}. A few nonparametric detectors for multivariate datastreams have been proposed in the literature \cite{boracchi2018quanttree, gretton2012kernel}, but most of these address change-detection in a \emph{one-shot} scheme by performing independent statistical tests over fixed-sized batches of data. Thus, these algorithms do not leverage the whole datastream and usually perform worse than their online counterparts \cite{Ross2011}.

We present \emph{QuantTree Exponentially Weighted Moving Average (QT-EWMA)}, a nonparametric online change-detection algorithm that can effectively monitor multivariate datastreams while controlling the frequency of false alarms, namely detections that do not correspond to any distribution change. As in statistical hypothesis testing, having a certain number of false alarms is unavoidable in change detection. In particular, any change-detection algorithm is characterized by a trade-off between the frequency of false alarms and the detection power. In many applications, including our previous example of industrial process monitoring~\cite{popescu2021basic}, promptly detecting changes is crucial (e.g., to avoid a failure). Still, any false alarm might trigger a costly intervention (e.g., the replacement of functioning machinery). Therefore, one typically sets a false alarm probability compatible with the resources allocated for these interventions and implements a change-detection algorithm that minimizes the detection delay subject to this bound on false alarms~\cite{Ross2011}. 
Moreover, controlling false alarms enables a fair comparison between the detection power of different solutions. Unfortunately, most online change-detection algorithms for multivariate datastreams from the literature, especially the nonparametric ones, cannot control false alarms effectively.

Typically, a change-detection algorithm has three main ingredients: \emph{i)} a \emph{model} $\widehat\phi_0$ of the initial distribution $\phi_0$ to be fitted on a training set, \emph{ii)} a \emph{statistic} $T$, based on $\widehat\phi_0$, that yields a known response when data are drawn from $\phi_0$, and \emph{iii)} a \emph{decision rule} to analyze the values of $T$ and report changes. Typically, the decision rule consists in comparing $T$ with a threshold $h$ defined to yield the desired false alarm probability. 
One-shot detectors, which analyze a fixed amount of data, have thresholds that do not depend on $t$ and are simply defined as quantiles of $T$, which can either be computed analytically \cite{lung2011robust} or by Monte Carlo simulations \cite{boracchi2018quanttree}. In contrast, the test statistic in online algorithms depends on $t$, as $T_t$ takes into account all the data points acquired until $t$. In this case, computing the thresholds is more complicated since one typically wants to control the \emph{Average Run Length} ($\arl{0}$), i.e. the expected time before raising a false alarm~\cite{basseville1993}. 



The proposed QT-EWMA is a novel online nonparametric change-detection algorithm for multivariate datastreams. QT-EWMA combines a \emph{QuantTree (QT)} histogram~\cite{boracchi2018quanttree}, used as a model $\widehat\phi_0$, and a novel online statistic $T_t$ based on \emph{Exponentially Weighted Moving Average (EWMA)}~\cite{roberts1959control}. In particular, by QT-EWMA we monitor the proportion of incoming samples falling in each bin of the histogram, and use this to build an efficient and practical online change-detection algorithm. The theoretical properties of QuantTree~\cite{boracchi2018quanttree} guarantee that QT-EWMA is completely nonparametric since the distribution of our statistic does not depend on $\phi_0$, hence its thresholds $\{h_t\}_t$ controlling the $\arl{0}$ can be set \emph{a priori}. Moreover, these thresholds guarantee by design a constant false alarm probability over time and, consequently, a fixed false alarm rate at any time instant during monitoring. Thus, QT-EWMA controls both $\arl{0}$ and false alarm rate.

We also introduce \emph{QT-EWMA-update}, a new change-detection algorithm based on QT-EWMA that enables online monitoring even when the training set is extremely small, e.g. in concept-drift adaptation~\cite{lu2018learning} when the change-detection algorithm has to be re-configured after a detection. In QT-EWMA-update we use new samples to update the bin probabilities of our initial QuantTree histogram, as long as no change is detected. This update improves the model $\widehat\phi_0$, thus increasing the detection power. Our updating procedure is compatible with the computational requirements of online monitoring schemes, and the distribution of the QT-EWMA-update statistic is also independent from $\phi_0$, enabling the computation of thresholds controlling the $\arl{0}$ through the same procedure as in QT-EWMA. Hence, QT-EWMA-update overcomes a major limitation characterizing several online and nonparametric change-detection algorithms, i.e., requiring a large training set to fit $\widehat\phi_0$ before monitoring. This is particularly useful when acquiring stationary data from $\phi_0$ is difficult or costly. 
Our main contributions are:
\begin{itemize}
    \item We present QT-EWMA, an online nonparametric change-detection algorithm for multivariate datastreams based on a novel EWMA statistic (Section \ref{subsec:ewma}).
    \item We prove that the bin probabilities of QuantTree histograms follow a Dirichlet distribution, and this allows us to compute the thresholds $\{h_t\}_t$ of QT-EWMA by an efficient Monte Carlo scheme. These thresholds enable controlling the $\arl{0}$ and false alarm rates for any $\phi_0$ (Section \ref{subsec:thresholds}).
    \item We propose QT-EWMA-update, which enables monitoring when the training set is extremely small by updating the QuantTree histogram online (Section \ref{sec:update}).
    \item We propose two simple yet theoretically sound procedures to extend a generic one-shot detector to monitor datastreams controlling the $\arl{0}$ (Section~\ref{sec:alternatives}), which we employ as baselines in our experiments.
\end{itemize}
Our experiments, performed on both synthetic and real-world datastreams, show that QT-EWMA controls the $\arl{0}$ better than the baselines and \emph{Scan-B}~\cite{li2015m}, a competing algorithm based on a \emph{Maximum Mean Discrepancy (MMD)} statistic~\cite{gretton2012kernel}, regardless of the training set size. Our results also show that QT-EWMA operates at the expected false alarm rate, which Scan-B does not guarantee. Moreover, QT-EWMA achieves similar or lower detection delays than Scan-B, especially on real-world datastreams. Most importantly, our QT-EWMA-update achieves significantly better detection performance compared to all the nonparametric alternatives when the training sets are extremely small. Our code and the thresholds of QT-EWMA and QT-EWMA-update are available at: \textcolor{blue}{\texttt{\url{https://boracchi.faculty.polimi.it/Projects}}}

This paper extends our previous work \cite{frittoli2021change}, where we introduced QT-EWMA. The major original contribution of this paper is QT-EWMA-update, which enables monitoring when an extremely small training set is provided. Moreover, we extend the results presented in \cite{boracchi2018quanttree} by proving that the bin probability vector of a QuantTree histogram is a realization of a Dirichlet random vector with known parameters, and this allows us to derive a very efficient Monte Carlo scheme to compute thresholds, 
reducing the runtime of the simulations by 25\% compared to \cite{frittoli2021change}.

The rest of the paper is organized as follows: in Section~\ref{sec:related} we survey the change-detection literature, focusing on methods operating on multivariate datastreams, and in Section~\ref{sec:problem} we provide a formal definition of the online change-detection problem. In Sections \ref{sec:solution} and \ref{sec:update} we introduce the QT-EWMA and QT-EWMA-update algorithms, respectively, and our procedure to compute thresholds controlling the $\arl{0}$. In Section~\ref{sec:alternatives} we illustrate how to extend one-shot change detectors to monitor datastream controlling the $\arl{0}$, and discuss the theoretical guarantees and limitations of these approaches. In Section~\ref{sec:complexity} we show that the computational complexity and memory requirements of QT-EWMA and QT-EWMA-update favorably compare to those of the alternative solutions and finally in Section~\ref{sec:experiments} we demonstrate the effectiveness of our solutions by testing it on both synthetic and real-world datastreams. 

\section{Related Work}\label{sec:related}


Most change-detection algorithms in the literature employ models and statistics designed to analyze univariate datastreams \cite{hawkins2003changepoint,Ross2011,roberts1959control}. The vast majority of these methods lack straightforward extensions to multivariate data, especially those leveraging nonparametric statistics based on ranks \cite{Ross2011}. Change detection in multivariate datastreams has often been addressed in \emph{multi-stream} (\emph{multi-channel}) settings, i.e., by separately analyzing each component of the datastream \cite{xie2013sequential,fellouris2017multichannel,sun2022quickest}. However, the hypotheses underpinning multi-stream monitoring are fundamentally different from those of change-detection in multivariate datastreams. In fact, \cite{xie2013sequential,fellouris2017multichannel,sun2022quickest} assume that the components of the input vector are generated by a 1-dimensional random variables, and ignore correlations among them. Moreover, in multi-stream settings, changes typically affect the distribution of a subset of these random variables \cite{xie2013sequential}, while, in multivariate settings, more general distribution changes are considered~\cite{carrera2018generating}, including subtle changes in the correlation between components that are hard to detect by multi-stream analysis. 

\begin{table}[t!]
\caption{Properties of the most relevant change-detection algorithms designed for multivariate datastream monitoring.}
\label{tab:properties}
\centering
\resizebox{\columnwidth}{!}{  
    \begin{tabular}{cc|c|c|c|c}
    \toprule
    & algorithm & nonpar. & online & control $\arl{0}$ & update\\
    \midrule
    \multirow{3}{*}{\rotatebox{90}{Gauss}} 
    & Hotelling CPM~\cite{zamba2006multivariate} & & \checkmark & \checkmark & \checkmark \\
    & SS-CPD~\cite{xie2020sequential} & & \checkmark & \checkmark & \\
    & SPLL~\cite{kuncheva2011change} & semipar. & with mod. & with mod. & \\
    \midrule
    \multirow{2}{*}{\rotatebox{90}{PCA}} 
    & PCA-SPLL~\cite{kuncheva2013pca} & \checkmark & & & \\
    & PCA-CD~\cite{qahtan2015pca} & \checkmark & \checkmark & & \\
    \midrule
    & Martingale~\cite{ho2005martingale,mozafari2011precise} & \checkmark & \checkmark & & \\
    \midrule
    \multirow{3}{*}{\rotatebox{90}{MMD}}
    & Scan-B~\cite{li2015m} & \checkmark & \checkmark & \checkmark & \checkmark \\
    & NEWMA~\cite{keriven2020newma} & \checkmark & \checkmark & iff $\phi_0$ known & \\
    & NTK-MMD~\cite{cheng2021neural} & \checkmark & \checkmark & & \\
    \midrule
    \multirow{4}{*}{\rotatebox{90}{Hist.}} 
    & BG-CuSum~\cite{lau2018binning} & \checkmark & \checkmark & iff $\phi_0$ known & \\
    & QuantTree~\cite{boracchi2018quanttree} & \checkmark & with mod. & with mod. & \\
    & QT-EWMA & \checkmark & \checkmark & \checkmark & \\
    & QT-EWMA-update & \checkmark & \checkmark & \checkmark & \checkmark \\
    \bottomrule
    \end{tabular}
}
\end{table}

Many change-detection algorithms specifically designed for multivariate datastreams are parametric: two remarkable examples are the \emph{Change Point Model (CPM)} \cite{zamba2006multivariate} based on the Hotelling test statistic \cite{hotelling1951generalized}, and \emph{Sequential Subspace Change Point Detection (SS-CPD)}~\cite{xie2020sequential}. Both methods perform online monitoring while controlling the $\arl{0}$, but rely on the hypothesis that $\phi_0$ is Gaussian. A popular approach to handle multivariate datastreams is to reduce the data dimension by computing the likelihood of the observations with respect to a Gaussian \cite{tartakovsky2006novel} or Gaussian mixture model $\widehat\phi_0$ \cite{kuncheva2011change,alippi2015change}, which is fitted on a training set and therefore is quite flexible in modeling $\phi_0$. As in \cite{kuncheva2011change}, we call these methods \emph{semiparametric}. The main limitation of parametric and semiparametric methods is the implicit assumption that $\phi_0$ belongs to a known family of probability distributions, which typically does not hold in real-world datastreams. Some nonparametric approaches reduce the data dimension by \emph{Principal Component Analysis (PCA)} \cite{kuncheva2013pca,qahtan2015pca}, or by a \emph{strangeness measure} \cite{ho2005martingale,mozafari2011precise} to monitor a univariate datastream, e.g. by Martingale-based permutation tests \cite{ho2005martingale}. However, none of these methods based on dimensionality reduction can be set to maintain the target $\arl{0}$.

The Maximum Mean Discrepancy (MMD) is a nonparametric statistic that was originally introduced for hypothesis testing \cite{gretton2012kernel}, and has recently been employed for online change detection \cite{li2015m,keriven2020newma,cheng2021neural} following a sliding-window approach. Usually, these methods do not fit a model $\widehat\phi_0$, but compare the new observations directly to the training set, which has to be stored during monitoring~\cite{li2015m}. Among these methods, Scan-B \cite{li2015m} is the only one where the $\arl{0}$ can be set before deployment for any unknown distribution $\phi_0$. However, Scan-B has a unique threshold, i.e. $h_t\equiv h$, defined by the asymptotic behavior of the $\arl{0}$ when $h\to\infty$ \cite{li2015m}, which does not guarantee an accurate control of the $\arl{0}$ and the false alarm rate, as we show in our experiments. \emph{NEWMA} \cite{keriven2020newma} detects changes by analyzing the relation between two EWMA statistics based on MMD having different forgetting factors. Unfortunately, thresholds controlling the $\arl{0}$ can be set only when the analytical expression of $\phi_0$ is known \cite{keriven2020newma}, which limits the applicability of NEWMA. \emph{Neural Tangent Kernel MMD (NTK-MMD)}~\cite{cheng2021neural} approximates the MMD statistic by training a neural network on samples from $\phi_0$, to reduce the computational and memory overhead in online testing. In this case, the thresholds are computed by training multiple networks with different training/validation splits, and then bootstrapping over validation data, a procedure that does not control the $\arl{0}$. A major limitation of algorithms based on MMD is that they require a large amount of reference data \cite{li2015m,cheng2021neural}: our experiments show that Scan-B \cite{li2015m} yields poor performance when the training set is small, even though the algorithm includes the incoming samples into the reference data, thus updating over time. In contrast, QT-EWMA does not require such large training sets, and QT-EWMA-update yields even lower detection delays by incrementally updating $\widehat\phi_0$.

Histograms are very flexible nonparametric models to describe $\phi_0$ \cite{boracchi2017uniform}. A remarkable example is QuantTree \cite{boracchi2018quanttree}, which adaptively defines a histogram over a training set drawn from $\phi_0$. QuantTree histograms have been employed in a one-shot change-detection test \cite{boracchi2018quanttree}, which cannot be directly used in online settings, leveraging a nonparametric statistic to assess whether a single batch of test data follows $\phi_0$ or not. 
Another change-detection algorithm based on histograms is the \emph{Binned Generalized Cumulative Sum (BG-CuSum)}~\cite{lau2018binning}, which can operate online controlling the $\arl{0}$. However, this algorithm has been tested only on univariate datastreams, and it is infeasible to extend to multivariate data because the number of bins scales exponentially with the data dimension. Moreover, it requires to know the cumulative function of $\phi_0$, or an accurate approximation, to enable controlling the $\arl{0}$. Thus, when $\phi_0$ is unknown, BG-CuSum requires a huge training set, especially when the data dimension is high \cite{lau2018binning}. The proposed QT-EWMA and QT-EWMA-update overcome all these limitations, enabling to control the $\arl{0}$ regardless of $\phi_0$ and the data dimension. Moreover, QT-EWMA-update is specifically designed to operate when the training set is small.

Table \ref{tab:properties} summarizes the main properties of the most relevant change-detection algorithms designed to monitor multivariate datastreams. In particular, we consider the following properties: being nonparametric, being executed online, controlling the $\arl{0}$, and being able to incrementally update the model using the incoming data. To the best of our knowledge, Scan-B~\cite{li2015m} is the only nonparametric and online change-detection algorithm from the literature in which the target $\arl{0}$ can be set independently of $\phi_0$. As we show in Section~\ref{sec:alternatives}, one-shot change-detection methods such as QuantTree~\cite{boracchi2018quanttree} and \emph{Semi-Parametric Log-Likelihood (SPLL)}~\cite{kuncheva2011change} can be modified to operate online while controlling the $\arl{0}$. Other methods that control the $\arl{0}$ are either parametric~\cite{zamba2006multivariate,xie2020sequential}, or require the analytical expression of $\phi_0$~\cite{keriven2020newma,lau2018binning}.

QT-EWMA-update shares some similarity with \emph{incremental learning} methods \cite{schlimmer1986incremental}, where streaming data samples are used to improve previously learned models \cite{he2011incremental}. In incremental learning and \emph{learning in non-stationary environments} literature, the model to be improved is typically a classifier \cite{ditzler2012incremental,alippi2013just}, hence a certain amount of supervised data is required. In contrast, we consider an unsupervised setting in which we do not know whether the incoming samples follow $\phi_0$ or not, and we incrementally update a model $\widehat\phi_0$ using streaming data, similarly to other online change-detection algorithms \cite{Ross2011,zamba2006multivariate}.

We remark that all the models and statistics described here might not be suitable to detect distribution changes in high-dimensional data such as signals or images. This is primarily due to the fact that models such as Gaussian mixtures \cite{kuncheva2011change} and histograms \cite{boracchi2018quanttree} cannot describe complicated structures, and in many statistics the computational overhead increases with the data dimension \cite{li2015m,lau2018binning}. Moreover, the higher the data dimension, the harder it is to detect distribution changes, an effect known as \emph{detectability loss} \cite{alippi2015change}. For this reason, high-dimensional data samples typically undergo a feature-extraction procedure to reduce their dimension before being analyzed by any change-detection algorithm. This is a standard procedure that has been followed to prepare the Credit Card Fraud Detection dataset~\cite{dal2017credit} and the INSECTS dataset~\cite{souza2020challenges}.

\section{Problem Formulation}
\label{sec:problem}
We address the online change-detection problem in a virtually unlimited multivariate datastream $x_1,x_2\ldots\in\mathbb{R}^d$. We assume that, as long as there are no changes, all the data samples are i.i.d. realizations of a random variable having unknown distribution $\phi_0$. In the case of time series, this hypothesis is typically met after some pre-processing~\cite{Ross2011}. We define the \emph{change point} $\tau$ as the unknown time instant when a change $\phi_0 \to \phi_1$ takes place:
\begin{equation}
    x_t\sim\begin{cases}
    \phi_0\;\; \text{if}\;\; t<\tau\\
    \phi_1\;\; \text{if}\;\; t\geq \tau
    \end{cases}.
    \label{eq:changepoint}
\end{equation}
We assume that both $\phi_0$ and $\phi_1 \neq \phi_0$ are unknown, and that a training set $TR$ containing $N$ realizations of $\phi_0$ is provided to fit $\widehat\phi_0$, which in our case is a QuantTree histogram \cite{boracchi2018quanttree}. After fitting $\widehat\phi_0$, an online change-detection algorithm assesses, for each new incoming sample $x_t$, whether the sequence $\{x_1,\ldots x_t\}$ contains a change point. Typically, a statistic $T_t$ based on $\widehat\phi_0$ is computed at each incoming $x_t$, then a decision rule is applied. Usually, the rule consists in controlling whether $T_t > h_t$ for a certain threshold $h_t$, and the detection time $t^*$ is defined as the first time instant when there is enough statistical evidence to claim that the datastream $\{x_1,\ldots x_{t^*}\}$ contains a change point, namely:
\begin{equation}
    t^* = \min\{t:T_t>h_t\}.
    \label{eq:detectiontime}
\end{equation}

As in any statistical test, the sequence of thresholds $\{h_t\}_t$ employed in change detection should be defined to control the probability of having a false alarm, namely a detection on data drawn from $\phi_0$. In online settings, we measure the amount of false alarms by the Average Run Length \cite{basseville1993}, defined as $\arl{0} = \mathbb{E}_{\phi_0}[t^*]$, where the expectation is taken assuming that the whole datastream is drawn from $\phi_0$. Thus, the $\arl{0}$ is the average time before a false alarm. Ideally, the \emph{target} $\arl{0}$ of an online change-detection method should be set \emph{a priori}, similarly to Type I error probability in hypothesis testing. The goal is to detect a distribution change as soon as possible, i.e., to minimize the \emph{detection delay} $t^*-\tau$, while \emph{controlling the} $\arl{0}$, i.e. having an \emph{empirical} $\arl{0}$ that approaches the target $\arl{0}$ set before monitoring. We remark that controlling the $\arl{0}$ also provides an upper bound on the expected detection delay.

When $TR$ is small, the model $\widehat\phi_0$ is typically inaccurate and the change-detection algorithm yields high detection delays. In this case, the algorithm should be able to incrementally update $\widehat\phi_0$ using new samples, as in \cite{Ross2011,zamba2006multivariate}, thus improving the detection performance.

\begin{algorithm}[t!]
\centering
\caption{QT-EWMA}
\label{alg:qt_ewma}
\begin{algorithmic}[1]
\Require datastream $x_1,x_2,\ldots$, target probabilities $\{\pi_j\}_{j=1}^{K}$, thresholds $\{h_t\}_t$, $TR$
\Ensure detection flag $\texttt{ChangeDetected}$, detection time $t^*$
\State $\texttt{ChangeDetected}\gets \text{False},\quad t^* \gets \infty$\;
\State estimate QT histogram $\{(S_j,\pi_j)\}_{j=1}^{K}$ from $TR$ and define $\{\tilde\pi_j\}$ as in \eqref{eq:approx}\label{line:qt}
\State $Z_{j,0}\gets\tilde\pi_j\;\forall j=1,\ldots,K$\;\label{line:start_ewma}
\For{$t=1,\ldots$}
    \State $y_{j,t} \gets \mathds{1}(x_t\in S_j)$\;\label{line:yj}
    \State $Z_{j,t} \gets (1-\lambda) Z_{j,t-1} + \lambda y_{j,t},\quad j=1\ldots,K$\;\label{line:end_ewma}
    \State $T_t \gets \sum_{j=1}^{K} (Z_{j,t}-\tilde\pi_j)^2 / \tilde\pi_j$\;\label{line:pearson_ewma}
    \If{$T_t > h_t$}\label{line:start_det}
        \State $\texttt{ChangeDetected}\gets \text{True}, \quad t^* \gets t$\;
        \State \textbf{break};
    \EndIf
\EndFor
\State \textbf{return} $\texttt{ChangeDetected}, t^*$\;\label{line:end_det}
\end{algorithmic}
\end{algorithm}

\section{QuantTree Exponentially Weighted Moving Average}
\label{sec:solution}
Here we introduce the QT-EWMA algorithm (Section \ref{subsec:ewma}) and illustrate the procedure we follow to compute its thresholds controlling the $\arl{0}$ (Section \ref{subsec:thresholds}). 

\subsection{The QT-EWMA Algorithm}
\label{subsec:ewma}
We propose QT-EWMA (Algorithm \ref{alg:qt_ewma}) to extend to online monitoring the QuantTree algorithm \cite{boracchi2018quanttree}, which was originally designed for one-shot change detection. QuantTree models $\phi_0$ by a histogram made of $K$ bins $\{S_j\}_{j=1}^{K}$ constructed by splitting $\mathbb{R}^d$ along random directions. The splits are defined so that each bin $S_j$ contains $\pi_jN$ samples from the training set $TR$, where $\{\pi_j\}_{j=1}^{K}$ is a given set of target probabilities. QuantTree histograms can model both univariate and multivariate distributions and, most importantly, enable nonparametric monitoring. In fact, the distribution of any statistic defined by the number of test samples falling in each bin $S_j$ of a QuantTree histogram does not depend on $\phi_0$ nor on the data dimension $d$, as demonstrated in \cite{boracchi2018quanttree}. Further details on QuantTree -- including how to define the bins when $TR$ cannot be exactly split to match the target probabilities -- can be found in \cite{boracchi2018quanttree}.


Here we define a novel online statistic $T_t$ to monitor the proportion of samples falling in each bin of a QuantTree histogram constructed over $TR$ (line \ref{line:qt}). In particular, when a new sample $x_t$ is acquired, we define $K$ binary statistics from the indicator functions of each bin $S_j$, namely
\begin{equation}\label{eq:y_j}
y_{j,t} = \mathds{1}(x_t\in S_j), \quad  j=1,\dots,K,
\end{equation}
to track in which bin $x_t$ falls. Denoting the true bin probabilities $p_j=\mathbb{P}_{\phi_0}(S_j)$, namely, the probability of a point sampled from $\phi_0$ to belong to $S_j$, we have that
\begin{equation}
    \label{eq:moments_pij}
    \begin{aligned}
    \mathbb{E}_{\phi_0}[y_{j,t}] = p_j, \; j = 1,\ldots, K,
    \end{aligned}
\end{equation}
where the expected value $\mathbb{E}_{\phi_0}$ is computed under the assumption that $x_t\sim\phi_0$. Since $\phi_0$ is unknown, so are the bin probabilities $(p_1,\ldots,p_K)$, which are a realization of a random vector \cite{boracchi2018quanttree} and can be approximated by $\tilde\pi_j\approx p_j$, where $\tilde\pi_1,\ldots,\tilde\pi_K$ are defined as:
\begin{equation}\label{eq:approx}
    \tilde\pi_j:=\frac{\pi_j N}{N+1}, j < K \; \text{and} \; \tilde\pi_K:=\frac{\pi_K N + 1}{N+1}.
\end{equation}
After evaluating the statistics $y_{j,t}$ for the incoming sample $x_t$ (line \ref{line:yj}), we compute the EWMA statistic \cite{roberts1959control} $Z_{j,t}$ (line \ref{line:end_ewma}), to monitor the proportion of data in $S_j$, for $j\in\{1,\ldots,K\}$:
\begin{equation}
    Z_{j,t} = (1-\lambda) Z_{j,t-1} + \lambda y_{j,t} \quad \text{ where } \quad Z_{j,0} = \tilde\pi_j\,.
    \label{eq:qtewma}
\end{equation}
Finally, we define the QT-EWMA change-detection statistic: 
\begin{equation}
    T_t = \sum_{j=1}^{K} \dfrac{(Z_{j,t}-\tilde\pi_j)^2}{\tilde\pi_j}\,,
    \label{eq:sequentialewma}
\end{equation}
which is similar to the Pearson statistic \cite{lehmann2006testing}. In fact, $T_t$ measures the overall difference between the proportion of samples $x_1,\ldots,x_t$ falling in each bin $S_j$, represented by $Z_{j,t}$, and $\tilde\pi_j$, which represent their estimated expected values under $\phi_0$. This difference naturally increases when $t>\tau$ as a consequence of a change $\phi_0\to\phi_1$ since this modifies the probability of some bin $S_j$. The QT-EWMA statistic is computed at each incoming sample (line \ref{line:pearson_ewma}) and then compared against the corresponding threshold $h_t$ to detect changes (line \ref{line:start_det}).


The distribution of any statistic defined over a QuantTree histogram does not depend on $\phi_0$ nor on $d$, thus QT-EWMA is a nonparametric change-detection algorithm. This claim is substantiated by the theoretical results in \cite{boracchi2018quanttree}, 
which we extend here by fully characterizing the probability distribution of $(p_1,\ldots,p_K)$:
\begin{proposition}\label{prop:dirichlet}
Let $\{S_j\}_{j=1}^K$ be a partitioning built by the QuantTree algorithm with target probabilities $\{\pi_j\}_{j=1}^K$ on a training set $TR\sim\phi_0$ of size $N$. Then, the bin probability vector $(p_1,\ldots,p_K)$ is drawn from the Dirichlet distribution:
\begin{equation}\label{eq:dirichlet}
    (p_1,\ldots,p_K)\sim\mathcal{D}\big(\pi_1 N,\pi_2N,\ldots,\pi_K N+1\big).
\end{equation}
\end{proposition}
\begin{proof}
We leverage the result in \cite{frigyik2010introduction} linking the Dirichlet distribution to the \emph{stick-breaking process}. In particular, the stick-breaking process generates a sequence of $K$ random variables $q_1, \dots, q_K$ as
\begin{equation}\label{eq:stick_breaking}
    q_j = \prod_{k=1}^{j-1}(1-\tilde q_k) \cdot \tilde q_j, \; j<K, \quad q_K = 1-\sum_{j=1}^{K-1}q_j,
\end{equation}
where $\tilde q_j$ for $j=1,\dots,K-1$ are defined as
\begin{equation}\label{eq:qtilde}
    \tilde q_j \sim \text{Beta}\bigg(\gamma_{j}, \sum_{k=j+1}^K \gamma_{j}\bigg),
\end{equation}
and $\gamma_1, \dots,\gamma_K$ are the parameters that define the stick-breaking process. In \cite{frigyik2010introduction} it has been shown that
\begin{equation}
    (q_1,\ldots,q_K)\sim\mathcal{D}\big(\gamma_1,\ldots,\gamma_K\big).
\end{equation}
To prove the proposition it is enough to show that there exists a specific configuration of $\gamma_j$ such that the bin probabilities $p_j$ of a QuantTree histogram can be expressed as $q_j$ in \eqref{eq:stick_breaking}. To this purpose, we recall the result in \cite{boracchi2018quanttree} where it has been shown that $p_j$ can be written as
\begin{equation}\label{eq:ptilde}
    p_j = \prod_{k=1}^{j-1}(1-\tilde p_k) \cdot \tilde p_j, \; j<K, \quad p_K = 1-\sum_{j=1}^{K-1}p_j,
\end{equation}
where $\tilde p_j$ are independent and follow Beta distributions:
\begin{equation}\label{eq:beta}
    \tilde p_j \sim \text{Beta}\bigg(\pi_j N, \bigg(1-\sum_{k=1}^j \pi_k\bigg)N + 1 \bigg)\;j=1,\ldots,K-1.
\end{equation}
Now, we only need to find a suitable choice of $\gamma_1,\ldots,\gamma_K$ to express the $\tilde p_j$ as the $\tilde q_j$ in \eqref{eq:qtilde}. If we define $\gamma_j= \pi_jN$ for $j<K$ as in \eqref{eq:beta} and $\gamma_K = \pi_KN +1$, we obtain that:
\begin{equation}\label{eq:gamma}
\begin{aligned}
    \sum_{k=j+1}^K\gamma_k = \sum_{k=j+1}^{K}\pi_kN + 1 = \left(1-\sum_{k=1}^j\pi_k\right)N + 1,
\end{aligned}
\end{equation}
where the last equality follows from $\sum_{j=1}^K\pi_j=1$. Equation \eqref{eq:gamma} ensures the correspondence between $\tilde p_j$ in \eqref{eq:beta} and $\tilde q_j$ in \eqref{eq:qtilde}, which implies the thesis.
\end{proof}

Proposition \ref{prop:dirichlet} means that, whenever we construct a QuantTree over a training set, we are partitioning $\mathbb{R}^d$ into $K$ bins with probabilities $(p_1,\ldots,p_K)$ drawn from the Dirichlet distribution in \eqref{eq:dirichlet}. Since the expected value of the $j$-th component of a random vector drawn from the Dirichlet distribution $\mathcal{D}(\gamma_1,\ldots,\gamma_K)$ is $\gamma_j/\sum_{k=1}^K\gamma_k$, by simple algebraic manipulation we have that $\mathbb{E}_{\phi_0}[p_j] = \tilde\pi_j$, where $\tilde\pi_j$ are defined as in \eqref{eq:approx}. Therefore, the values $\tilde\pi_1,\ldots,\tilde\pi_K$ can be used as estimates of the bin probabilities $p_1,\ldots,p_K$. Moreover, from the property of the Dirichlet distribution we have that $\text{var}[p_j]\to0$ when $N\to\infty$, thus $\tilde\pi_j$ is a good estimate of $p_j$. We remark that the statistics employed in QuantTree \cite{boracchi2018quanttree} estimate the bin probability $p_j$ by its target value $\pi_j$ since it is assumed that a large training set is provided, and $\tilde\pi_j\to\pi_j$ as $N\to\infty$ by definition \eqref{eq:approx}. Here we also consider cases where $N$ is small, thus in the QT-EWMA statistics \eqref{eq:qtewma} and \eqref{eq:sequentialewma} we employ $\tilde\pi_j$, which is a more accurate estimate of $p_j$.

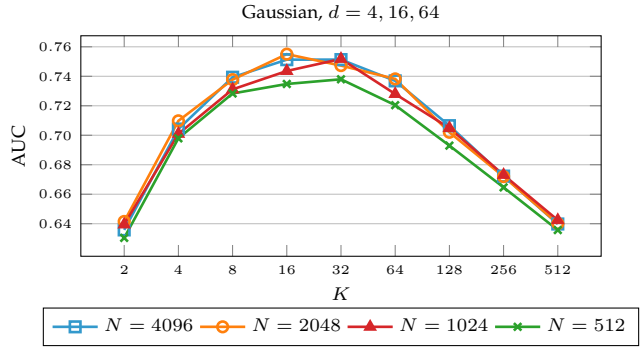
\begin{figure}[t!]
    \centering
    \begin{tikzpicture}
    \begin{axis}[
        style_bins,
        xlabel={$K$},
        ylabel={AUC},
        title={Gaussian, $d=4,16,64$},
        ytick = {0.64,0.66,...,0.8},
        ymajorgrids=true,
        yticklabel style = {font = \tiny,
            /pgf/number format,
    	    fixed,
            fixed zerofill,
    	    precision=2},
        legend style={at={(0.5,-0.32)},
        	font=\scriptsize,
        	legend columns=-1,fill=none,draw=black,anchor=center,align=center,
        	},
        ]
        
        \addplot [
        style_qtewma,
        ]
        table [
            x expr=\coordindex,
            y=N4096,
        ]{data/AUC_K.txt};
        \addlegendentry{$N=4096$}
        
        \addplot [
        style_qt,
        ]
        table [
            x expr=\coordindex,
            y=N2048,
        ]{data/AUC_K.txt};
        \addlegendentry{$N=2048$}
        
        \addplot [
        style_stop1024,
        ]
        table [
            x expr=\coordindex,
            y=N1024,
        ]{data/AUC_K.txt};
        \addlegendentry{$N=1024$}
        
        \addplot [
        style_stop512,
        ]
        table [
            x expr=\coordindex,
            y=N512,
        ]{data/AUC_K.txt};
        \addlegendentry{$N=512$}
        
    \end{axis}

\end{tikzpicture}
    \vspace{-3mm}
    \caption{Detection power of QT-EWMA on Gaussian datastreams with different training set size $N$ when varying the number of bins $K$. The results, which are averaged over $d=4,16,64$, show that histograms with a small $K$ cannot describe $\phi_0$ accurately and yield low detection performance. Also setting a very large $K$ harms detection performance since, at a fixed $N$, increasing $K$ yields inaccurate estimates $\{\tilde\pi_j\}_j$.}
    \label{fig:auc_K}
\end{figure}

\noindent\textbf{Impact of the choice of $K$.} The number of bins $K$ of the QuantTree histogram is a fundamental parameter that influences the change-detection performance of QT-EWMA. To analyze the impact of the choice of $K$, we test QT-EWMA with $K=2,4,8,\dots,512$. In particular, we compute the QT-EWMA statistic over 5000 stationary Gaussian datastreams and 5000 datastreams containing a change point at $\tau=500$, and we measure the detection power by the \emph{Area Under the ROC Curve (AUC)} given by the statistic values at $t=1000$, namely $T_{1000}$. The distribution changes $\phi_0\rightarrow\phi_1$ consist in random roto-translations of $\phi_0$ generated by CCM~\cite{carrera2018generating} (see Section \ref{subsec:datasets}). Since the detection performance depends also on the training set size $N$, we employ different values of $N=512,1024,2048,4096$. 

In Fig. \ref{fig:auc_K} we report the average results obtained on datastreams with $d=4,16,64$. Setting $K=2,4$ yields low AUC because histograms having such few bins cannot describe $\phi_0$ well. Increasing the number of bins ($K=128,256,512$) while keeping $N$ fixed increases the variance of the bin probabilities $p_j$ due to the properties of the Dirichlet distribution \eqref{eq:dirichlet}, harming the detection performance. Intermediate values -- especially $K=16,32$ for the considered values of $N$ -- yield the best results, thus we in our experiments we select $K=32$ as in \cite{boracchi2018quanttree}. As expected, the detection performance increases with $N$ since $\text{var}[p_j]\to0$ when $N\to\infty$. However, the improvement is substantial only when increasing $N$ from $512$ to $1024$, while using a larger $N$ yields a marginal improvement, see Fig. \ref{fig:auc_K}.

\subsection{Computing Thresholds to Control the ARL$_0$}
\label{subsec:thresholds}
In online monitoring, the thresholds $\{h_t\}_t$ should guarantee the target $\arl{0} = \mathbb E_{\phi_0}[t^*]$, where $t^*$ is the detection time, as defined in \eqref{eq:changepoint}. Thanks to the properties of QuantTree~\cite{boracchi2018quanttree}, the distribution of any statistic based on QuantTree, including $T_t$, does not depend on $\phi_0$ nor on $d$. Therefore, the QT-EWMA thresholds $\{h_t\}_t$ defined to yield the target $\arl{0}$ will only depend on the EWMA parameter $\lambda$, the target bin probabilities $\{\pi_j\}_{j=1}^K$, and the training set size $N$. Following~\cite{margavio1995alarm}, we define $\{h_t\}_t$ to guarantee a fixed false alarm probability $\alpha$ at each time instant $t$. This implies that the detection time $t^*$ under $\phi_0$ is a Geometric random variable with parameter $\alpha$ \cite{margavio1995alarm}, hence its expected value is
\begin{equation}
     \arl{0} = \mathbb E_{\phi_0}[t^*] = \dfrac{1}{\alpha}.
     \label{eq:arl_alpha}
\end{equation}
To this purpose, as noted in \cite{margavio1995alarm}, the thresholds $\{h_t\}_t$ must satisfy the following condition:
\begin{equation}
    \mathbb P_{\phi_0}(T_t>h_t \;| \;T_k\leq h_k\; \forall k < t) = \alpha\quad \forall t\geq1.
    \label{eq:cond_prob}
\end{equation}
Since it is infeasible to exactly compute the conditional probabilities in \eqref{eq:cond_prob}, we resort to Monte Carlo simulations as in \cite{Ross2011}. Leveraging Proposition \ref{prop:dirichlet}, we simulate the construction of a QuantTree histogram on a training set $TR\sim\phi_0$ of size $N$ by drawing its bin probabilities $(p_1,\ldots,p_K)$ from the Dirichlet distribution \eqref{eq:dirichlet}. Then, for each probability vector, we simulate the binary statistics $(y_{1,t},\ldots,y_{K,t})$ in \eqref{eq:y_j} of a stationary datastream of length $L=5000$ by drawing them from the following multinomial distribution $\mathcal{M}$: 
\begin{equation}\label{eq:multinomial}
    (y_{1,t},\ldots,y_{K,t})\sim\mathcal{M}(p_1,\ldots,p_K).
\end{equation}
Then, we use these values $\{(y_{1,t},\ldots,y_{K,t})\}_{t=1}^{5000}$ to compute the QT-EWMA statistics $\{T_t\}_{t=1}^{5000}$ by \eqref{eq:qtewma}--\eqref{eq:sequentialewma}. To compute the thresholds $\{h_t\}_t$ yielding the desired $\arl{0}$, we repeat the procedure above 1,000,000 times, and define $h_1$ as the empirical $(1-\alpha)$-quantile of all the values of $T_1$, where $\alpha = 1/\arl{0}$ as in \eqref{eq:arl_alpha}. Similarly, we define $h_t$ with $t>1$ as the $(1-\alpha)$-quantiles of the values $T_t$, using only those sequences $\{(y_{1,k},\ldots,y_{K,k})\}_{k=1}^t$ whose statistics $T_k$ have never exceeded any of the previous thresholds $h_k$ for $k=1,\ldots,t-1$. Computing the thresholds $\{h_t\}_t$ in this way guarantees that, for each time $t$, the empirical quantiles of $T_t$ are conditioned to $T_k\leq h_k\; \forall k < t$, which in turn implies \eqref{eq:cond_prob}, hence the target $\arl{0}$ is preserved \cite{margavio1995alarm}.

We compute the thresholds $h_t$ for $t=1,\ldots,5000$ and then fit a polynomial in powers of $1/t$ to these values that returns $h_t$ for a given $t$, as suggested in \cite{Ross2011}. This allows to both estimate $h_t$ for $t>5000$ and to improve the estimates $\{h_t\}_{t=1}^{5000}$ by leveraging correlation among thresholds. In our code we provide the polynomial expressions of the thresholds maintaining $\arl{0}=500,1000,2000,5000,10000,20000$, which can be very useful to control false alarms in high-throughput applications.

This procedure based on Proposition \ref{prop:dirichlet} is substantially more efficient than that presented in~\cite{frittoli2021change}, where we computed the QT-EWMA statistics $\{T_t\}_{t=1}^{5000}$ from synthetic univariate Gaussian datastreams, i.e. $\phi_0=\mathcal{N}(0,1)$ after constructing a QuantTree histogram on a synthetic training set $TR\sim\phi_0$. Directly generating the bin probabilities $(p_1,\ldots,p_K)$ from the Dirichlet distribution \eqref{eq:dirichlet} and the sequences $\{(y_{1,t},\ldots,y_{K,t})\}_{t=1}^{5000}$ from the multinomial distribution \eqref{eq:multinomial} replaces the construction of a QuantTree histogram on each training set and the computation of the binary statistics $(y_{1,t},\ldots,y_{K,t})$ \eqref{eq:y_j} for each synthetic datastream, reducing by 25\% the average runtime of the Monte Carlo simulations compared to \cite{frittoli2021change}.

\noindent\textbf{Control over False Alarm Rates.} An important consequence of setting a constant false alarm probability in \eqref{eq:cond_prob} is that our thresholds can also control the false alarm rate at any time instant $t$. In fact, being $t^*$ a Geometric random variable \cite{margavio1995alarm} with parameter $\alpha$, the probability of having a false alarm before $t$ corresponds to the following geometric sum:
\begin{equation}\label{eq:geosum}
    \mathbb{P}_{\phi_0}(t^*\leq t) = \sum_{k=1}^t \alpha(1-\alpha)^{k-1} = 1 - (1-\alpha)^t.
\end{equation}
This property enables us to assess the control of false alarms on datastreams containing a change point at $\tau$ by computing the proportion of datastreams in which $t^*\leq\tau$. This can then be compared to the target false positive rate in \eqref{eq:geosum}, which depends on the target $\arl{0}$ (see Section \ref{subsec:figuresofmerit}).

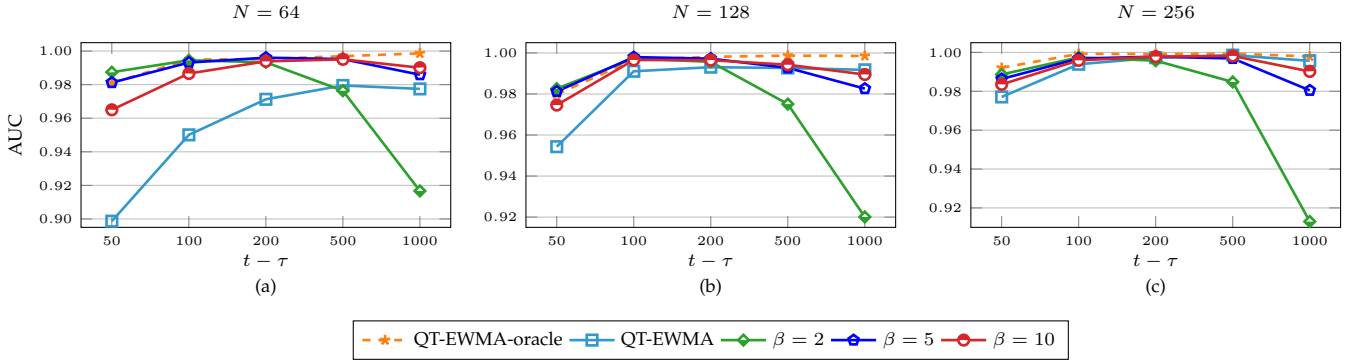
\begin{figure*}[t!]
    \centering
    \begin{tikzpicture}
\begin{groupplot}[ 
    group style={group size=3 by 1,}, 
    width=0.27\textwidth,
    height=0.135\textwidth,
    title style = {font=\scriptsize,
    },
    scale only axis,
    ymajorgrids,
    xlabel style = {anchor = near ticklabel, 
        	at = {(0.5,-0.1)},font=\scriptsize},
    ylabel style = {font=\scriptsize},
    xticklabel style = {font = \tiny},
    xtick = {1,2,3,4,5},
    xticklabels = {$50$,$100$,$200$,$500$,$1000$}, 
    yticklabel style = {font = \tiny,
    /pgf/number format,
    	    fixed,
            fixed zerofill,
    	    precision=2},
]
%
\nextgroupplot[
        xlabel={$t-\tau$},
        ylabel={AUC},
        title={$N=64$},
        ymin = 0.895,
        ymax = 1.005,
        ytick = {0.9,0.92,...,1},
        ymajorgrids=true,
        /pgf/number format/.cd,
        scale only axis,scaled ticks=false,]
\addplot [style_truep,]
table [
    x=delta,
    y=truep,
]{data/AUC_64.txt};

\addplot [style_qtewma,]
table [
    x=delta,
    y=qtewma,
]{data/AUC_64.txt};

\addplot [style_beta2,]
table [
    x=delta,
    y=beta2,
]{data/AUC_64.txt};

\addplot [style_beta5,]
table [
    x=delta,
    y=beta5,
]{data/AUC_64.txt};

\addplot [style_beta10,]
table [
    x=delta,
    y=beta10,
]{data/AUC_64.txt};

\nextgroupplot[
        xlabel={$t-\tau$},
        title={$N=128$},
        ymin = 0.915,
        ymax = 1.005,
        ytick = {0.9,0.92,...,1},
        ymajorgrids=true,
        /pgf/number format/.cd,
        scale only axis,scaled ticks=false,]
\addplot [style_truep,]
table [
    x=delta,
    y=truep,
]{data/AUC_128.txt};

\addplot [style_qtewma,]
table [
    x=delta,
    y=qtewma,
]{data/AUC_128.txt};

\addplot [style_beta2,]
table [
    x=delta,
    y=beta2,
]{data/AUC_128.txt};

\addplot [style_beta5,]
table [
    x=delta,
    y=beta5,
]{data/AUC_128.txt};

\addplot [style_beta10,]
table [
    x=delta,
    y=beta10,
]{data/AUC_128.txt};

\nextgroupplot[
        xlabel={$t-\tau$},
        title={$N=256$},
        ymin = 0.91,
        ymax = 1.005,
        ytick = {0.9,0.92,...,1},
        ymajorgrids=true,
        legend style={at={($(0,0)+(0cm,0cm)$)},legend columns=-1,fill=none,draw=black,anchor=center,align=center, font=\scriptsize, legend to name = grouplegend,},
        /pgf/number format/.cd,
        scale only axis,scaled ticks=false,]
\addplot [style_truep,]
table [
    x=delta,
    y=truep,
]{data/AUC_256.txt};
\addlegendentry{QT-EWMA-oracle}

\addplot [style_qtewma,]
table [
    x=delta,
    y=qtewma,
]{data/AUC_256.txt};
\addlegendentry{QT-EWMA}

\addplot [style_beta2,]
table [
    x=delta,
    y=beta2,
]{data/AUC_256.txt};
\addlegendentry{$\beta=2$}

\addplot [style_beta5,]
table [
    x=delta,
    y=beta5,
]{data/AUC_256.txt};
\addlegendentry{$\beta=5$}

\addplot [style_beta10,]
table [
    x=delta,
    y=beta10,
]{data/AUC_256.txt};
\addlegendentry{$\beta=10$}

\end{groupplot}
\node[align=center] at ($(group c1r1.south west)!.5!(group c1r1.south east) + (0cm,-0.8cm)$) {\scriptsize(a)};
\node[align=center] at ($(group c2r1.south west)!.5!(group c2r1.south east) + (0cm,-0.8cm)$) {\scriptsize(b)};
\node[align=center] at ($(group c3r1.south west)!.5!(group c3r1.south east) + (0cm,-0.8cm)$) {\scriptsize(c)};
\node at ($(group c2r1.south west)!.5!(group c2r1.south east) + (0cm,-1.5cm)$) {\ref{grouplegend}}; 
\end{tikzpicture}%
    \vspace{-0.5cm}
    \caption{Detection power of QT-EWMA-update ($\beta=2,5,10$) compared to QT-EWMA ($\beta=\infty$) and the oracle QT-EWMA over univariate datastreams containing a change point at $\tau=1000$, setting $N=64,128,256$. In particular, we compute the AUC of the statistic $T_t$ at different times $t>\tau$. QT-EWMA-update outperforms QT-EWMA right after the change, especially when $N=64$. However, the performance of QT-EWMA-update decreases over time since it updates the bin probabilities using $x_t\sim\phi_1$ when $t>\tau$, and this is very apparent when the updating speed is high ($\beta=2$).
    }
    \label{fig:auc}
\end{figure*}

\section{Updating the QuantTree Histogram}\label{sec:update}
Here we present QT-EWMA-update (Section \ref{subsec:update}), evaluate the impact of the updating speed on the detection performance (Section \ref{subsec:discussion}), and discuss stopping the update to avoid including post-change samples (Section \ref{subsec:stopping}).

\subsection{The QT-EWMA-update Algorithm}
\label{subsec:update}
In QT-EWMA we model the distribution $\phi_0$ by means of a QuantTree histogram~\cite{boracchi2018quanttree} constructed on a training set $TR$ of size $N$. 
Then, during monitoring, we compute the statistic $T_t$ \eqref{eq:sequentialewma} to compare the proportion of samples falling in each bin $S_j$ with the estimated bin probabilities $\tilde\pi_j$. Being $\phi_0$ unknown, we approximate the true bin probabilities $p_j=\mathbb{P}_{\phi_0}(S_j)$ by $\tilde\pi_j$, which is reasonable since $\mathbb{E}_{\phi_0}[p_j] = \tilde\pi_j$ and $\text{var}[p_j]\to0$ as $N\to\infty$ thanks to Proposition \ref{prop:dirichlet}. However, when $N$ is small, the variance of $p_j$ is high, thus $\{\tilde\pi_j\}_{j=1}^K$ yield inaccurate estimates of the true bin probabilities $\{p_j\}_{j=1}^K$, which harms the detection performance. 

To overcome such limitation in online settings, we propose to update the model $\widehat\phi_0$ every time a new observation $x_t$ arrives, which increases the detection power as in \cite{zamba2006multivariate,Ross2011}. In particular, we present \emph{QT-EWMA-update}, where we replace each $\tilde\pi_j$ in \eqref{eq:sequentialewma} with an estimate $\hat p_{j,t}$ of the bin probability $p_j$ that is incrementally updated when a new observation $x_t$ becomes available, as long as no changes are detected. We define $\hat p_{j,t}$ as:
\begin{equation}\label{eq:update}
    \hat p_{j,0} = \tilde\pi_j, \quad \hat p_{j,t} = (1-\omega_t)\hat p_{j,t-1} + \omega_t y_{j,t} \quad t>0,
\end{equation}
where $y_{j,t}=\mathds{1}(x_t\in S_j)$, $\omega_t=1/\beta(N+t)$ is a parameter representing the \emph{updating speed} as it regulates the weight of the latest sample in the average, and $\beta\geq1$ is a tuning parameter. 
We remark that all the quantities involved in our QT-EWMA-update statistic, including $\hat p_{j,t}$, are computed from a QuantTree histogram, thus the distribution of the statistic $T_t$ does not depend on $\phi_0$ nor on $d$~\cite{boracchi2018quanttree}. Therefore, we can compute the thresholds of QT-EWMA-update for a given $\beta$ by the same Monte Carlo procedure presented in Section \ref{subsec:thresholds}, guaranteeing the control of the $\arl{0}$.

\subsection{The Role of the Updating Speed}\label{subsec:discussion}

The parameter $\beta$ allows tuning the updating speed $\omega_t$ of QT-EWMA-update, which has a crucial impact on the detection performance. Setting $\beta=1$ guarantees that $\hat p_{j,t}\to p_j$ when $t\to\infty$, as long as $x_t\sim\phi_0$, since \eqref{eq:update} becomes the cumulative average of $y_{j,t}$, whose expected value is $\mathbb{E}_{\phi_0}[y_{j,t}]=p_j$ by definition. However, samples $x_t\sim\phi_1$ acquired after the change $\tau$ introduce a severe bias that harms the detection performance. Therefore, we propose to set $\beta>1$, as this reduces the contribution of the most recent samples when updating $\hat p_{j,t}$. Setting $\beta>1$ slightly biases the estimate $\hat p_{j,t}$ in stationary conditions, but turns out to be very beneficial in terms of detection delay, as shown in our experiments. We remark that QT-EWMA corresponds to the case $\beta=\infty$ since $\hat p_{j,t}\equiv\tilde\pi_j$. 

To illustrate the trade-off regulated by $\beta$, we perform a simple experiment on univariate datastreams, setting $\phi_0=\mathcal{U}(0,1)$ and $\phi_1=\mathcal{N}(0.5,0.5)$. We generate 1000 univariate training sets from $\phi_0$, 500 stationary univariate datastreams of length $L=2000$ from $\phi_0$, and 500 datastreams with initial distribution $\phi_0$ containing a change point at $\tau=1000$.

We monitor each datastream by the QT-EWMA-update algorithm setting $\beta=2,5,10$, and QT-EWMA. We measure the detection power of these algorithms by the AUC of the statistics $T_t$ computed at different times $t$ after the change such that $t-\tau=50,100,200,500,1000$. Since in this case we set $\phi_0=\mathcal{U}(0,1)$, we can easily compute the bin probabilities $p_j$ of each QuantTree histogram, so we also test the ``oracle'' QT-EWMA algorithm, which uses $p_j$ instead of $\tilde\pi_j$ in \eqref{eq:qtewma} and \eqref{eq:sequentialewma}, and that is never updated. The oracle is an upper bound in terms of AUC, as it uses the analytical expression of $\phi_0$. 

In Fig. \ref{fig:auc} we show the results of this experiment, using training set size $N=64,128,256$. We observe that the QT-EWMA algorithm steadily improves its detection power when more samples drawn from $\phi_1$ are considered. We also observe that the oracle QT-EWMA yields much better results than QT-EWMA when the training set is small, while the gap reduces when $N$ is sufficiently large, as $\tilde\pi_j$ becomes an accurate estimate of $p_j$. In all cases, QT-EWMA-update yields a higher detection power compared to QT-EWMA right after $\tau$, but shows a substantial decrease of the AUC over time due to the fact that the $\hat p_{j,t}$ are updated using samples from $\phi_1$. Using a larger $\beta$ mitigates this effect, at the cost of slightly reducing the detection power right after $\tau$. In our experiments (Section \ref{sec:experiments}) we set $\beta=5$ as it yields the best detection performance in Fig. \ref{fig:auc}.

\subsection{Stopping the Update}\label{subsec:stopping}
All in all, QT-EWMA-update outperforms QT-EWMA when the training set is small ($N=64,128$), but the update yields only a marginal advantage when $N=256$ because $\tilde\pi_j$ is already a good estimate of $p_j$. Hence, when a large training set is available, QT-EWMA is preferable since it avoids the risk of updating the model using samples from the post-change distribution. QT-EWMA-update is preferred when $N$ is small, but the update of $\widehat\phi_0$ should stop as soon as a sufficient number $S$ of samples have been acquired without detecting a change, i.e., when $N+t=S$. This allows to reduce the risk of updating using samples $x_t\sim\phi_1$ once the estimated bin probabilities $\hat p_{j,t}$ are sufficiently accurate. Since stopping the update does not change the the fact that the distribution of the statistic is independent from $\phi_0$ and $d$, which is guaranteed by the properties of QuantTree \cite{boracchi2018quanttree}, we compute thresholds controlling the $\arl{0}$ for given values of $\beta$ and $S$ using the same Monte Carlo scheme illustrated in Section \ref{subsec:thresholds}. The only difference with respect to QT-EWMA-update is that we update the bin probabilities $\hat p_{j,t}$ by \eqref{eq:update} only for $t<S-N$, using $\hat p_{j,S-N}$ when $t\geq S-N$.

\section{Online One-shot Change Detection} \label{sec:alternatives}
In this section we show how to adapt one-shot algorithms, namely statistical tests to assess whether a fixed amount of samples was generated by $\phi_0$, to online change detection controlling the $\arl{0}$. We focus on algorithms that operate batch-wise (Section \ref{subsec:batchwise}), and element-wise (Section \ref{subsec:reduction}).

\subsection{Datastream Monitoring by Batch-wise Detectors}\label{subsec:batchwise}
Several change-detection algorithms process the datastream in separate non-overlapping batches $W_t$ of $\nu$ samples:
\begin{equation}
    W_t = [x_{(t-1)\nu + 1}, \dots, x_{t\nu}].
\end{equation}
In particular, these algorithms compute for each incoming batch $W_t$ a test statistic $T^{\nu}(W_t)$ based on a model $\widehat\phi_0$ fit over $TR$. For example, in QuantTree \cite{boracchi2018quanttree} $\widehat\phi_0$ is a histogram, while SPLL \cite{kuncheva2011change} employs a Gaussian mixture. These algorithms detect a change as soon as $T^{\nu}(W_t) > h^{\nu}$, where the threshold $h^{\nu}$ does not depend on $t$ and is defined to control the false alarm probability over each batch $W_t$. In what follows we show how to set the threshold $h^{\nu}$ in batch-wise monitoring algorithms to maintain the target $\arl{0}$ in online change detection, leveraging the following results:
\begin{proposition}\label{prop:arl_batch}
Let $W_t$ be any batch of $\nu$ samples drawn from $\phi_0$ and let the detection threshold $h^{\nu}$ be such that 
\begin{equation}\label{eq:fpr}
\mathbb P_{\phi_0}(T^{\nu}(W_t) > h^{\nu}) = \alpha.
\end{equation}
Then, the monitoring scheme $T^{\nu}(W_t) > h^{\nu}$ yields $\arl{0}\geq \nu/\alpha$.
\end{proposition}
\begin{proof}
Reported in the supplementary material.
\end{proof}

Therefore, setting $\alpha=\nu/\arl{0}$, any batch-wise monitoring algorithm can be transformed into a conservative online change-detection algorithm, guaranteeing that the $\arl{0}$ is greater than or equal to the target. A slightly different result holds when the threshold $h^{\nu}$ is conditioned on $TR$, e.g. when $h^{\nu}$ is computed by bootstrap. The following Proposition shows that, in this case, setting $\alpha=\nu/\arl{0}$ guarantees that the $\arl{0}$ is equal to the target.

\begin{proposition}\label{prop:exact}
Let $W_t$ be any batch of $\nu$ samples drawn from $\phi_0$ and let the detection threshold $h^{\nu}$ be such that 
\begin{equation}\label{eq:fpr_bootstrap}
\mathbb P_{\phi_0}(T^{\nu}(W_t) > h^{\nu} \;|\; TR) = \alpha,
\end{equation}
Then, the monitoring scheme $T^{\nu}(W_t) > h^{\nu}$ yields $\arl{0}= \nu/\alpha$.
\end{proposition}
\begin{proof}
Reported in the supplementary material.
\end{proof}

Leveraging these results, we adapt two well-known batch-wise change-detection methods to monitor datastreams online while controlling the $\arl{0}$: QuantTree \cite{boracchi2018quanttree} and SPLL~\cite{kuncheva2011change}. The properties of QuantTree \cite{boracchi2018quanttree} guarantee that it is possible to set $h^{\nu}$ for \eqref{eq:fpr} to hold for the Pearson statistic~\cite{lehmann2006testing}, independently from $\phi_0$ and $TR$. Hence, Proposition~\ref{prop:arl_batch} allows to set a lower bound on the $\arl{0}$. In contrast, the distribution of the SPLL statistic, namely the log-likelihood, depends on $\phi_0$, so $h^{\nu}$ has to be computed by bootstrapping over a portion of $TR$ that was not used to fit $\widehat\phi_0$. In this case, the hypothesis of Proposition~\ref{prop:exact} holds since the false positive probability is conditioned on the provided $TR$, thus the online version of SPLL yields the target $\arl{0}$. The main drawback of this bootstrap procedure is that it requires a large $TR$ to fit $\widehat\phi_0$ and to compute $h^{\nu}$.

\subsection{Datastream Monitoring by Element-wise Detectors}\label{subsec:reduction}
As pointed out in Section \ref{sec:related}, an approach to change detection in multivariate datastreams consists in reducing the data dimension, constructing a univariate datastream that can be monitored by standard change-detection algorithms. Here we reduce the dimension of each incoming sample $x_t$ as in SPLL \cite{kuncheva2011change} by computing the log-likelihood $-\log(\widehat{\phi}_0(x_t))$, where $\widehat{\phi}_0$ is a Gaussian mixture model fit on the entire $TR$. Then, we monitor the resulting univariate sequence by a nonparametric online CPM \cite{Ross2011} leveraging the Lepage test statistic \cite{Lepage1971}. This algorithm, which we call SPLL-CPM, maintains the desired $\arl{0}$ thanks to the CPM, which controls the $\arl{0}$ on any univariate datastream \cite{Ross2011}.

\begin{table}[t!]
\caption{Computational complexity for each update of the statistic and memory requirement of QT-EWMA and QT-EWMA-update compared to the other methods, depending on the configuration (Section \ref{subsec:config}). 
}
\label{tab:complexity}
\centering
    \begin{tabular}{c|c|c}
        \toprule
        algorithm & complexity & memory\\
        \midrule
        QT-EWMA & $\mathcal{O}(K)$ & $K$\\
        QT-EWMA-update & $\mathcal{O}(K)$ & $2K$\\
        QuantTree \cite{boracchi2018quanttree} & $\mathcal{O}(K)$ & $K$\\
        SPLL \cite{kuncheva2011change} & $\mathcal{O}(md)$ & $1$\\
        SPLL-CPM & $\mathcal{O}(md + w\log w)$ & $w$\\
        Scan-B \cite{li2015m} & $\mathcal{O}(nBd)$ & $(n+1)Bd$\\
        \bottomrule
    \end{tabular}
\end{table}

\section{Computational Complexity}\label{sec:complexity}

Since efficiency is key in online monitoring \cite{Ross2011}, we analyze the computational complexity and memory requirements of QT-EWMA and QT-EWMA-update. We perform the same analysis on the online versions of QuantTree \cite{boracchi2018quanttree} and SPLL~\cite{kuncheva2011change} (Section \ref{subsec:batchwise}), SPLL-CPM (Section \ref{subsec:reduction}), and Scan-B~\cite{li2015m} (Section \ref{sec:related}). The results are summarized in Table \ref{tab:complexity}.

\begin{figure*}[t!]
    \centering
    \input{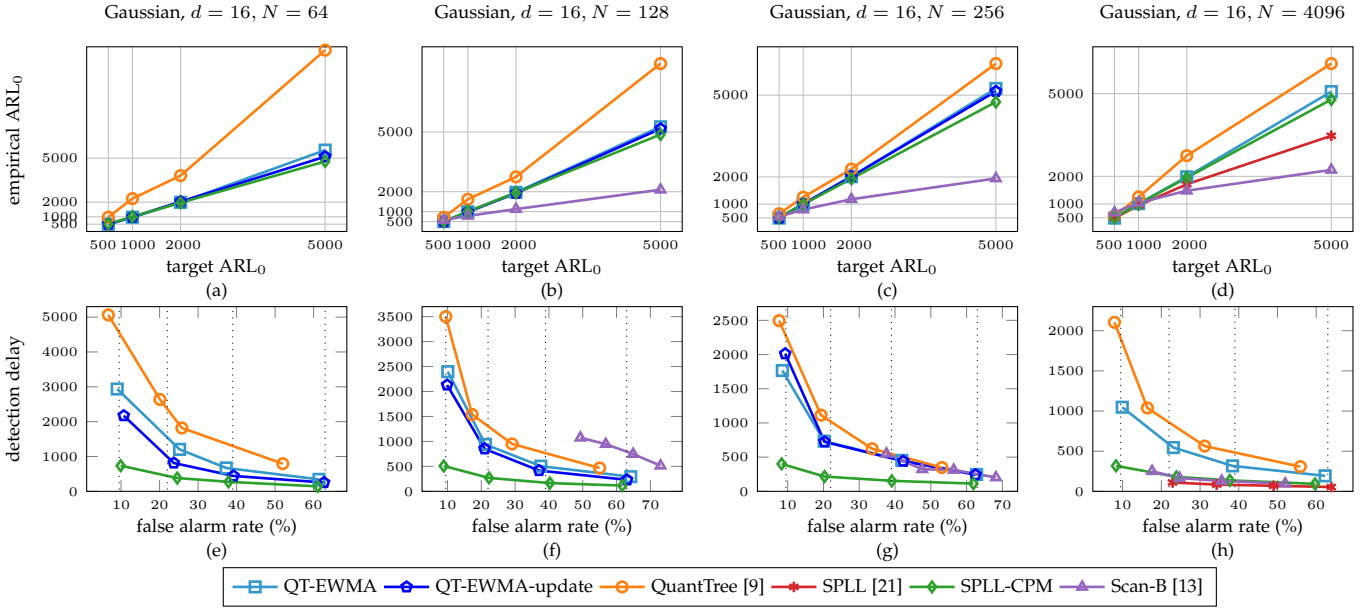}
    \vspace{-0.4cm}
    \caption{Experimental results over Gaussian datastreams ($d=16$). (a,b,c,d) show that the empirical $\arl{0}$ of QT-EWMA, QT-EWMA-update and SPLL-CPM approaches the target, while the other methods do not maintain the target $\arl{0}$. (e,f,g,h) show that, in terms of detection delay, the best-performing method is SPLL-CPM when using small training sets ($N=64,128,256$) and SPLL when using large training sets ($N=4096$). We observe that only QT-EWMA, QT-EWMA-update and SPLL-CPM achieve the target false alarm rates given by \eqref{eq:geosum}, which are represented by the vertical dotted lines.}
    \label{fig:gaussian16}
\end{figure*}

\noindent\textbf{QT-EWMA, QT-EWMA-update and QuantTree.} These algorithms are extremely efficient and require an amount of memory that is constant over time and does not depend on the data dimension $d$. Before monitoring, a QuantTree histogram is constructed, requiring to rank the training set $K$ times according to a specific component, resulting in $\mathcal{O}(KN\log N)$ operations~\cite{boracchi2018quanttree}, where $K$ is the number of bins and $N$ is the training set size. During monitoring, these three algorithms find the bin of the QuantTree histogram where each incoming sample $x_t$ falls, resulting in $\mathcal{O}(K)$ operations \cite{boracchi2018quanttree}. Then, QT-EWMA and QT-EWMA-update compute the test statistics \eqref{eq:y_j}, \eqref{eq:qtewma}, \eqref{eq:sequentialewma} at a constant overhead that falls within $\mathcal{O}(K)$. QT-EWMA-update also updates the bin probabilities of the QuantTree histogram by \eqref{eq:update}, requiring $K$ additional operations, which also fall within $\mathcal{O}(K)$. The QuantTree algorithm instead computes the Pearson statistic at the end of each batch, and this does not increase the order of computational complexity either, resulting in $\mathcal{O}(K)$ operations as in QT-EWMA and QT-EWMA-update. In terms of memory requirement, QT-EWMA only stores the $K$ values $Z_{j,t-1}, j = 1,\dots,K$ to compute \eqref{eq:qtewma} for each new sample $x_t$. QT-EWMA-update stores also the $K$ estimated bin probabilities $\hat p_{j,t-1}, j = 1,\dots,K$, hence it requires to store $2K$ values in total. Similarly to QT-EWMA, QuantTree stores only the proportion of points in the batch belonging to each of the $K$ bins to compute the Pearson statistic.

\noindent\textbf{SPLL and SPLL-CPM.} Both these algorithms are based on a Gaussian mixture model $\widehat{\phi}_0$ with $m$ components fitted on $TR$. In SPLL, the likelihood of an incoming batch $W_t$ is computed incrementally (before applying the logarithm) as the average likelihood of the samples $x_{(t-1)\nu + 1}, \dots, x_{t\nu}$, requiring $\mathcal{O}(md)$ operations per sample \cite{kuncheva2011change}. Hence, only 1 value has to be stored in memory, namely the likelihood computed in the previous step. In contrast, the SPLL-CPM algorithm leverages the CPM framework \cite{Ross2011} to monitor the stream of log-likelihood values $\{-\log(\widehat{\phi}_0(x_t))\}_t$. In particular, the Lepage test statistic \cite{Lepage1971} used in the CPM requires to sort the whole log-likelihood sequence obtained until time $t$, resulting in $\mathcal{O}(t\log t)$ operations on top of the $\mathcal{O}(md)$ operations required to compute $-\log(\widehat{\phi}_0(x_t))$. In this case, all the $t$ values of the log-likelihood sequence have to be processed and stored at each time $t$, thus the computational complexity and memory requirement steadily increase over time. Since this is not desirable in online settings, the ranks of older observations can be discretized and stored in a histogram, yielding an approximation of the Lepage statistic~\cite{Ross2011} using only the most recent $w$ samples.

\noindent\textbf{Scan-B.} The Scan-B algorithm \cite{li2015m} operates on sliding windows of size $B$, using $n$ windows sampled from $TR$ as a reference. For each incoming sample $x_t$, Scan-B updates $n$ Gram matrices by computing $B$ times the MMD statistic, resulting in $\mathcal{O}(nBd)$ operations \cite{keriven2020newma}. The $n$ reference windows and the current window have to be stored, yielding $(n+1)Bd$ values in memory \cite{keriven2020newma}. Thus, the computational and memory requirements of Scan-B increase with $d$.

\section{Experiments}\label{sec:experiments}
In this section we show that QT-EWMA and QT-EWMA-update can control the $\arl{0}$ and false alarm rates substantially better than competing methods, while achieving lower or comparable detection delays. We perform our experiments in two configurations: large ($N=4096$) and small training sets ($N=64,128,256$), to show the advantages of QT-EWMA-update when $N$ is small.


\begin{figure*}[t!]
    \centering
    \input{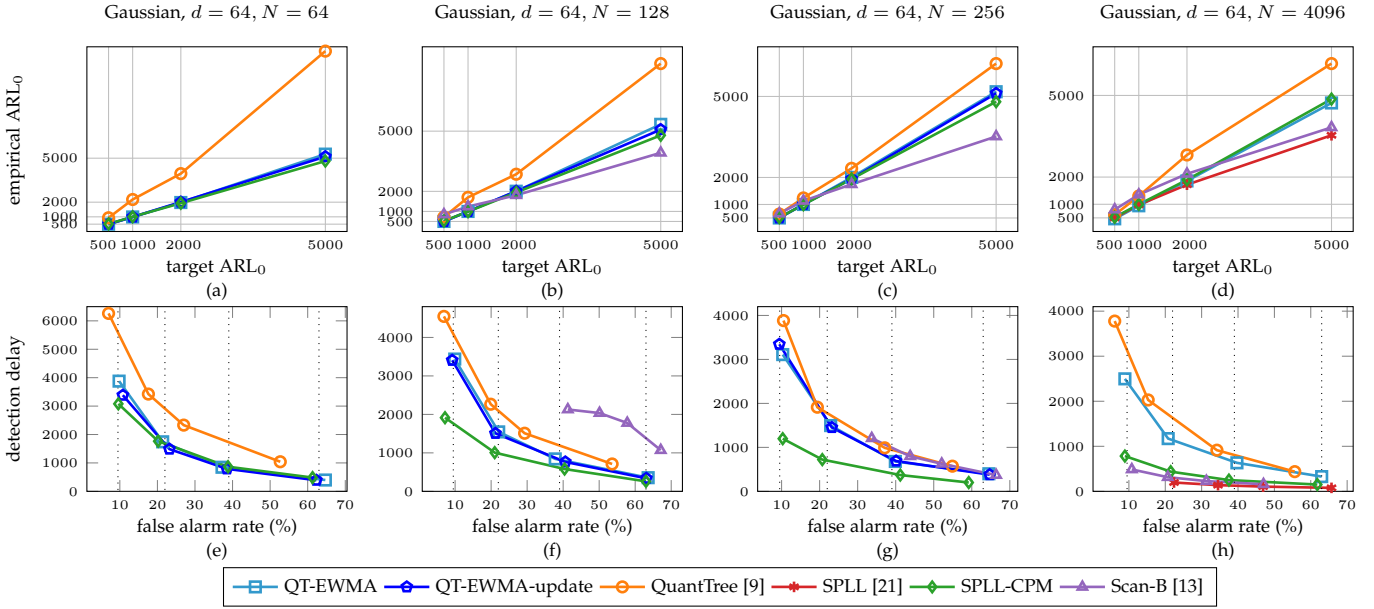}
    \vspace{-8mm}
    \caption{Experimental results over Gaussian datastreams ($d=64$). (a,b,c,d) show that the empirical $\arl{0}$ of QT-EWMA, QT-EWMA-update and SPLL-CPM approaches the target, while the other methods do not maintain the target $\arl{0}$. (e,f,g,h) show that, in terms of detection delay, the best-performing method is SPLL-CPM when using small training sets ($N=64,128,256$) and SPLL when using large training sets ($N=4096$). We observe that only QT-EWMA, QT-EWMA-update and SPLL-CPM achieve the target false alarm rates given by \eqref{eq:geosum}, which are represented in the plots by vertical dotted lines.}
    \label{fig:gaussian64}
\end{figure*}

\subsection{Considered Datasets}\label{subsec:datasets}
We simulate Gaussian datastreams of dimension $d=4,16,64$, choosing an initial Gaussian distribution $\phi_0$ with random mean and covariance matrix, and roto-translating $\phi_0$ to obtain the post-change distribution $\phi_1=\phi_0(Q\cdot+v)$. We randomly select the roto-translation parameters $Q$ and $v$ using the CCM framework \cite{carrera2018generating} to guarantee a symmetric Kullback-Leibler divergence $\text{sKL}(\phi_0,\phi_1)=0.5,1,1.5,2,2.5,3$. These settings are very useful to compare the detection performance at a different $d$ \cite{alippi2015change}. For brevity, here we report only the results on Gaussian data with $d=16,64$, while $d=4$ is in the supplementary material.

We also test on seven real-world multivariate datasets: Credit Card Fraud Detection (``credit'', $d=28$) from \cite{dal2017credit}, Sensorless Drive Diagnosis (``sensorless'', $d=48$), MiniBooNE particle identification (``particle'', $d=50$), Physicochemical Properties of Protein Ternary Structure (``protein'', $d=9$), El Ni\~no Southern Oscillation (``ni\~no'', $d=5$), and two of the Forest Covertype datasets (``spruce'' and ``lodgepole'', $d=10$) from the UCI Machine Learning Repository \cite{uci2019}. As in \cite{boracchi2018quanttree}, we standardize the datasets and sum to the samples of ``sensorless'', ``particle'', ``spruce'' and ``lodgepole'' an imperceptible Gaussian noise to avoid repeated values, which harm the construction of QuantTree histograms. We prepare datastreams by randomly sampling these datasets, whose distribution can be considered stationary, and we introduce a change by applying a shift of a random vector drawn from a standard $d$-dimensional Gaussian distribution, scaled by the total variance of the dataset, as in \cite{boracchi2018quanttree,kuncheva2013pca}. For brevity, we report only the average results over the ``UCI+credit'' datasets, while the results over individual datasets are in the supplementary material.

We also test on the INSECTS dataset \cite{souza2020challenges} ($d=33$), which contains features describing the wing-beat frequency of different species of flying insects, extracted from high-dimensional signals acquired by optical sensors. This dataset is meant as a classification benchmark for datastreams affected by concept drift. The dataset contains six concepts, each referring to data acquired under different environmental conditions affecting the insects' behavior. We assemble data from different concepts to form datastreams that include 30 types of realistic changes: we start sampling observations from one concept ($\phi_0$) and switch to another ($\phi_1$) introducing a change point. 

To make sure that training and test data do not have samples in common, we generate Gaussian training and test data from different seeds, and sample real-world datastreams after removing $TR$ from the datasets \cite{dal2017credit,uci2019,souza2020challenges}.

\begin{figure*}[t!]
    \centering
    \input{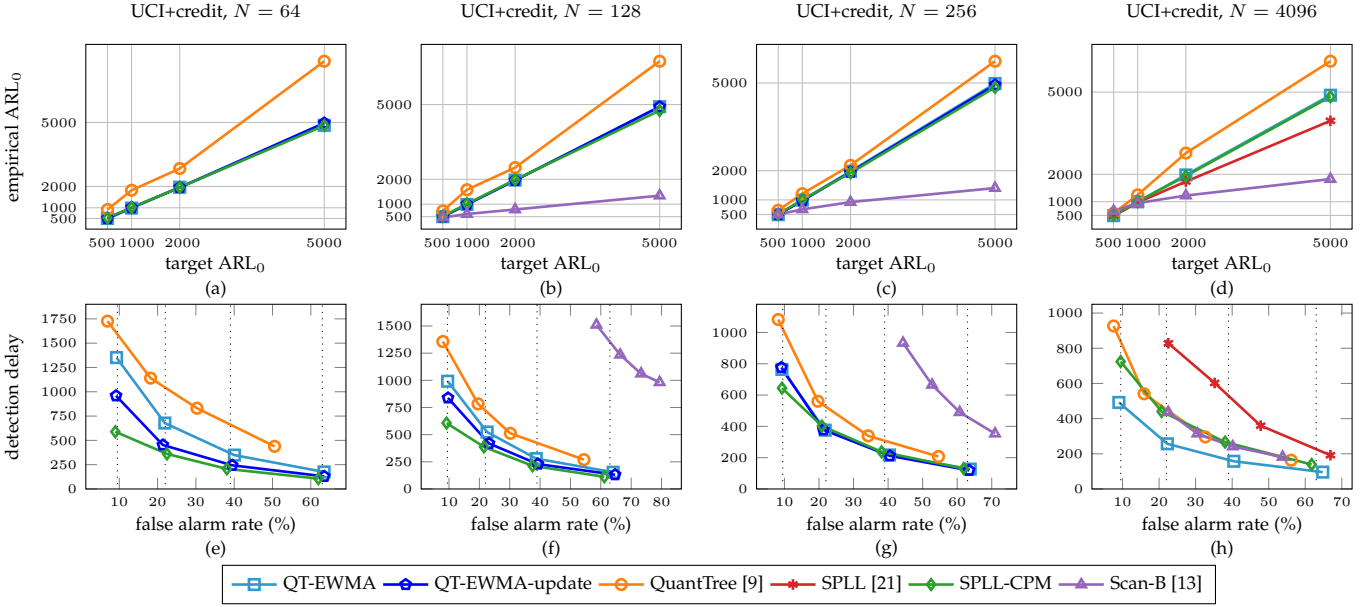}
    \vspace{-4mm}
    \caption{Experimental results averaged over the UCI+credit datasets \cite{uci2019,dal2017credit}. (a,b,c,d) show that the empirical $\arl{0}$ of QT-EWMA, QT-EWMA-update and SPLL-CPM approaches the target, while the other methods do not maintain the target $\arl{0}$. (e,f,g,h) show that, in terms of detection delay, the best-performing methods are QT-EWMA-update and SPLL-CPM when using small training sets ($N=64,128,256$) and QT-EWMA when using large training sets ($N=4096$). We observe that only QT-EWMA, QT-EWMA-update and SPLL-CPM achieve the target false alarm rates given by \eqref{eq:geosum}, which are represented in the plots by vertical dotted lines. 
    }
    \label{fig:uci_arl}
\end{figure*}

\subsection{Considered Methods}
\label{subsec:config}
To enable a fair comparison, we only consider change-detection methods where the target $\arl{0}$ can be set before monitoring, regardless of $\phi_0$. As shown in Section~\ref{sec:related}, the vast majority of the existing methods do not control the $\arl{0}$ or do so only when $\phi_0$ is known~\cite{keriven2020newma,xie2020sequential}, which is not guaranteed in general. For this reason, we compare QT-EWMA and QT-EWMA-update against QuantTree~\cite{boracchi2018quanttree}, SPLL~\cite{kuncheva2011change}, SPLL-CPM (described in Section~\ref{sec:alternatives}), and Scan-B~\cite{li2015m}, which is the only method from the literature where the $\arl{0}$ can be set independently on $\phi_0$. Here we illustrate the configuration of the considered methods.

\noindent\textbf{QT-EWMA, QT-EWMA-update and QuantTree.} In our experiments we adopt the standard configuration of QuantTree~\cite{boracchi2018quanttree}, with $K=32$ bins and uniform target probabilities $\pi_j=1/K$. In fact, \cite{boracchi2017uniform} shows that uniform histograms are very effective for change detection purposes. In QT-EWMA-update we set the parameter $\beta$, which regulates the updating speed in \eqref{eq:update}, to $\beta=5$, which yields the best results in our preliminary experiment (see Fig. \ref{fig:auc}). In Section \ref{subsec:discussion} we have shown that updating the QuantTree histogram is beneficial when the training set is extremely small, hence we test QT-EWMA-update with $N=64,128,256$. In QuantTree we set the batch size $\nu=32$ as in~\cite{boracchi2018quanttree}.

\noindent\textbf{SPLL and SPLL-CPM.} When monitoring datastreams sampled from Gaussian distributions and from the UCI+credit datasets, we set the number of components of the Gaussian mixture $\widehat\phi_0$ to $m=1$. To maximize the performance on the INSECTS dataset, which contains data from 6 different species of insects \cite{souza2020challenges}, we set $m=6$. In SPLL, we set the batch size $\nu=32$ as for QuantTree, and employ 1/4 of the training set to fit $\widehat\phi_0$ and the remaining samples to compute the threshold by bootstrap, as illustrated in Section \ref{subsec:batchwise}. Since the threshold computation for SPLL requires a relatively large amount of data, we only test SPLL with large training sets ($N=4096$). In SPLL-CPM we use the entire training set to fit $\widehat\phi_0$, since the CPM employed to monitor the log-likelihood does not require a training set~\cite{Ross2011}.

\noindent\textbf{Scan-B.} In all the experiments with large training sets ($N=4096$) we test Scan-B \cite{li2015m} in its standard configuration, with $n=5$ windows of size $B=100$. This configuration cannot be employed when the training set is extremely small since Scan-B requires $N\geq nB$ \cite{li2015m}. For this reason, we set $B=50$ when $N=256$ and $B=20$ when $N=128$, keeping $n=5$. All these configurations are among those suggested in \cite{li2015m}. Since no configurations with $B<20$ are reported in~\cite{li2015m}, we do not test Scan-B when $N=64$.

\subsection{Figures of Merit}
\label{subsec:figuresofmerit}
\noindent\textbf{Empirical $\arl{0}$.} To assess whether QT-EWMA and the other considered methods maintain the target $\arl{0}$, we compute the empirical $\arl{0}$ as the average time before raising a false alarm. In particular, we set the target $\arl{0}=500,1000,2000,5000$, and prepare 5000 datastreams of length $L=6 \cdot \arl{0}$. According to \eqref{eq:geosum}, the probability of having a detection in these stationary datastreams is $\mathbb{P}_{\phi_0}(t^*\leq L)\approx0.9975$. 

\noindent\textbf{Detection delay.} We also evaluate the average detection delay, i.e. $\arl{1} = \mathbb E_{\phi_1}[t^* - \tau]$, where the expectation is taken assuming that a change point $\tau$ is present \cite{basseville1993}. We run the considered methods configured with $\arl{0}=500,1000,2000,5000$ on 1000 datastreams of length $L=10000$, each containing a change point at $\tau=500$. We compute the empirical $\arl{1}$ as the average difference $t^*-\tau$ over these datastreams, excluding those yielding false alarms.

\noindent\textbf{False alarm rate.} To assess whether the desired false alarm probability is achieved, we compute the percentage of datastreams in which a detection occurs at $t^*<\tau$. Setting the the target $\arl{0}=500,1000,2000,5000$ should yield a false alarm in 63\%, 39\%, 22\% and 9.5\% of the datastreams \eqref{eq:geosum}.

\begin{figure*}[t!]
    \centering
    \input{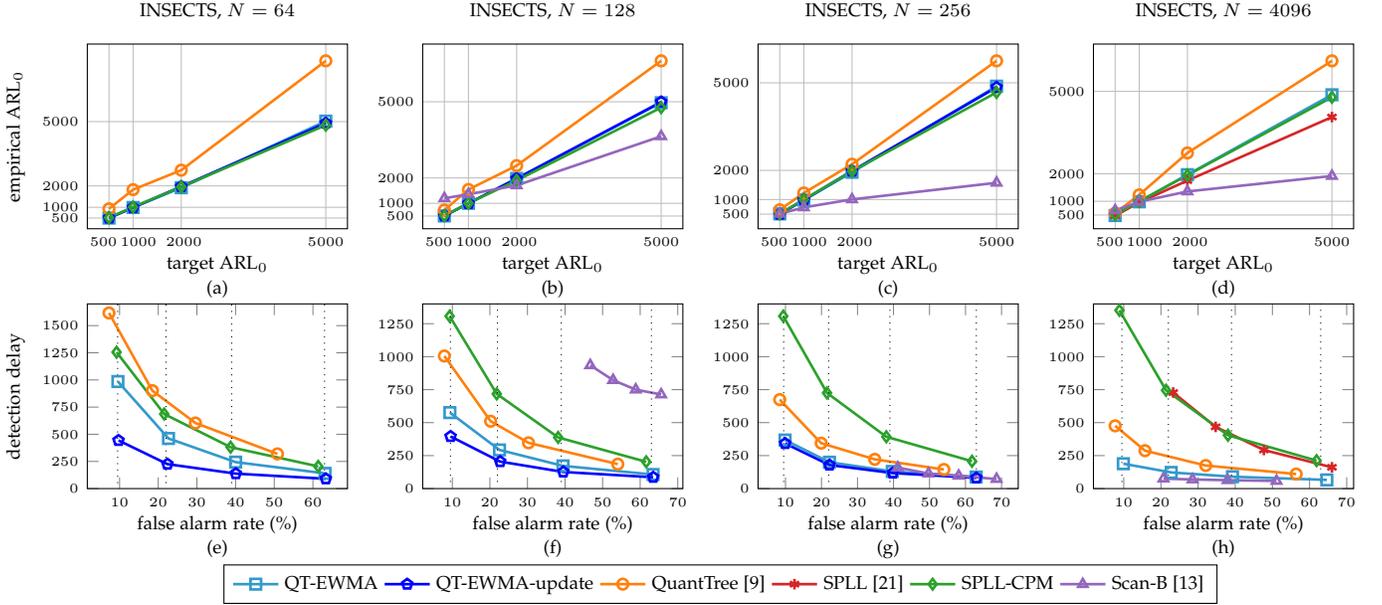}
    \vspace{-7mm}
    \caption{Experimental results averaged over the INSECTS dataset \cite{souza2020challenges}. (a,b,c,d) show that the empirical $\arl{0}$ of QT-EWMA, QT-EWMA-update and SPLL-CPM approaches the target, while the other methods do not maintain the target $\arl{0}$. (e,f,g,h) show that, in terms of detection delay, the best-performing method is QT-EWMA-update when using small training sets ($N=64,128,256$) and Scan-B when using large training sets ($N=4096$). We observe that only QT-EWMA, QT-EWMA-update and SPLL-CPM achieve the target false alarm rates given by \eqref{eq:geosum}, which are represented in the plots by vertical dotted lines.}
    \label{fig:insects_arl}
\end{figure*}

\subsection{Results and Discussion}
\label{subsec:results}
\noindent\textbf{Empirical $\arl{0}$.} Fig. \ref{fig:gaussian16} (a,b,c,d) and Fig. \ref{fig:gaussian64} (a,b,c,d) plot the empirical $\arl{0}$ achieved on Gaussian datastreams with $d=16$ and $d=64$, respectively, against the target $\arl{0}$. These plots show that QT-EWMA, QT-EWMA-update can control the $\arl{0}$ very accurately, independently from the data dimension $d$ and the training set size $N$. This can be seen from the fact that the lines are close to the diagonal (note that axis scales are different). The empirical $\arl{0}$ of QuantTree is higher than the target, and this is consistent with the statement of Proposition \ref{prop:arl_batch}. In contrast, the empirical $\arl{0}$ of Scan-B substantially departs from the target, in particular when the target $\arl{0}$ is large. The limitations of Scan-B in controlling the $\arl{0}$ are due to the fact that its threshold $h$ is defined by an asymptotic approximation of the $\arl{0}$ as $h\to\infty$ and $h/\sqrt{B}\to c$, where $c$ is a constant \cite{li2015m}. Therefore, a larger target $\arl{0}$ requires a larger threshold $h$ and, in principle, a larger window size $B$. However, increasing $B$ is infeasible because it would increase the computational and memory requirements (Table \ref{tab:complexity}), and also the training set size since $N\geq nB$. 
Despite the theoretical guarantees of Proposition~\ref{prop:exact}, we observe that also SPLL cannot maintain the target $\arl{0}$ accurately, and this is due to inaccurate estimate of its thresholds, which are computed by bootstrap over a limited training set. In contrast, SPLL-CPM accurately controls the $\arl{0}$ thanks to the properties of the CPM that monitors the log-likelihood~\cite{Ross2011}. Results obtained on the UCI+credit and INSECTS datasets (Fig. \ref{fig:uci_arl} (a,b,c,d) and Fig. \ref{fig:insects_arl} (a,b,c,d)) are consistent with those achieved on synthetic data.

\noindent\textbf{Detection delay vs false alarm rate.} We plot the average detection delay against the percentage of false alarms to assess the trade-off between these two quantities. Fig. \ref{fig:gaussian16} (e,f,g,h) and Fig. \ref{fig:gaussian64} (e,f,g,h) illustrate the performance on Gaussian datastreams with a change point at $\tau=500$ and dimension $d=16,64$, respectively.

In terms of detection delay, QT-EWMA-update is the best nonparametric method when the training set is small ($N=64,128,256$), being outperformed only by SPLL-CPM, which operates in ideal conditions since its parametric assumptions are met ($\phi_0$ is a Gaussian). For the same reason, when the training set is large ($N=4096$), both SPLL and SPLL-CPM outperform QT-EWMA. Also Scan-B outperforms QT-EWMA in these settings since statistics defined on histograms (such as that of QT-EWMA) are known to be less powerful than those based on MMD (such as that of Scan-B), as they perceive only changes affecting bin probabilities, and are totally blind to distribution changes inside each bin. Nevertheless, our experiments show that both QT-EWMA and QT-EWMA-update yield lower detection delays than Scan-B when the training set is small. All methods achieve higher detection delays as $d$ increases due to detectability loss \cite{alippi2015change}, 
which becomes apparent when we set the change magnitude to $\text{sKL}(\phi_0,\phi_1)=0.5,1,1.5,2,2.5,3$ in the CCM framework \cite{carrera2018generating}. In the supplementary material we show that the detection delays of all the considered methods decreases when the change magnitude increases. 


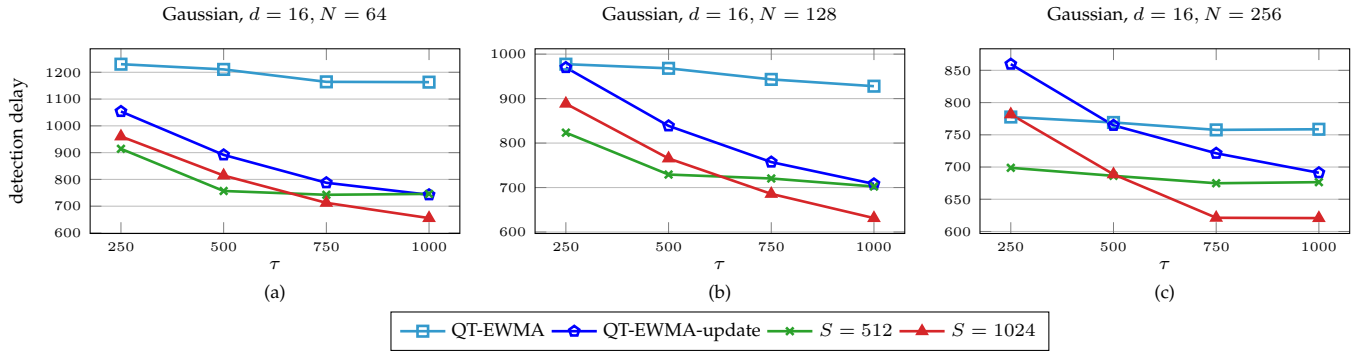
\begin{figure*}[t!]
    \centering
    \begin{tikzpicture}
\begin{groupplot}[ 
    group style={group size=3 by 1,}, 
    width=0.27\textwidth,
    height=0.135\textwidth,
    title style = {font=\scriptsize,
    },
    scale only axis,
    ymajorgrids,
    xlabel style = {anchor = near ticklabel, 
        	at = {(0.5,-0.1)},font=\scriptsize},
    ylabel style = {font=\scriptsize},
    xticklabel style = {font = \tiny},
    yticklabel style = {font = \tiny},
]
%
\nextgroupplot[
        title={Gaussian, $d=16,N=64$},
        ylabel={detection delay},
        xlabel={$\tau$},
        ytick = {600,700,...,1600},
        xtick = {250,500,750,1000},
        /pgf/number format/.cd,use comma,1000 sep={},scale only axis,scaled ticks=false,]

\addplot [style_qtewma,]
table [
    x=cp,
    y=qt_ewma,
]{data/stop_update_16_64_2000.txt};

\addplot [style_beta5,]
table [
    x=cp,
    y=update,
]{data/stop_update_16_64_2000.txt};

\addplot [style_stop512,]
table [
    x=cp,
    y=stop512,
]{data/stop_update_16_64_2000.txt};

\addplot [style_stop1024,]
table [
    x=cp,
    y=stop1024,
]{data/stop_update_16_64_2000.txt};

\nextgroupplot[
        title={Gaussian, $d=16,N=128$},
        xlabel={$\tau$},
        ytick = {500,600,...,1000},
        xtick = {250,500,750,1000},
        /pgf/number format/.cd,use comma,1000 sep={},scale only axis,scaled ticks=false,]

\addplot [style_qtewma,]
table [
    x=cp,
    y=qt_ewma,
]{data/stop_update_16_128_2000.txt};

\addplot [style_beta5,]
table [
    x=cp,
    y=update,
]{data/stop_update_16_128_2000.txt};

\addplot [style_stop512,]
table [
    x=cp,
    y=stop512,
]{data/stop_update_16_128_2000.txt};

\addplot [style_stop1024,]
table [
    x=cp,
    y=stop1024,
]{data/stop_update_16_128_2000.txt};

\nextgroupplot[
        title={Gaussian, $d=16,N=256$},
        xlabel={$\tau$},
        ytick = {500,550,...,1000},
        xtick = {250,500,750,1000},
        legend style={at={($(0,0)+(0cm,0cm)$)},legend columns=-1,fill=none,draw=black,anchor=center,align=center, font=\scriptsize, legend to name = grouplegend,},
        /pgf/number format/.cd,use comma,1000 sep={},scale only axis,scaled ticks=false,]

\addplot [style_qtewma,]
table [
    x=cp,
    y=qt_ewma,
]{data/stop_update_16_256_2000.txt};
\addlegendentry{QT-EWMA}

\addplot [style_beta5,]
table [
    x=cp,
    y=update,
]{data/stop_update_16_256_2000.txt};
\addlegendentry{QT-EWMA-update}

\addplot [style_stop512,]
table [
    x=cp,
    y=stop512,
]{data/stop_update_16_256_2000.txt};
\addlegendentry{$S=512$}

\addplot [style_stop1024,]
table [
    x=cp,
    y=stop1024,
]{data/stop_update_16_256_2000.txt};
\addlegendentry{$S=1024$}

\end{groupplot}
\node[align=center] at ($(group c1r1.south west)!.5!(group c1r1.south east) + (0cm,-0.8cm)$) {\scriptsize(a)};
\node[align=center] at ($(group c2r1.south west)!.5!(group c2r1.south east) + (0cm,-0.8cm)$) {\scriptsize(b)};
\node[align=center] at ($(group c3r1.south west)!.5!(group c3r1.south east) + (0cm,-0.8cm)$) {\scriptsize(c)};
\node at ($(group c2r1.south west)!.5!(group c2r1.south east) + (0cm,-1.3cm)$) {\ref{grouplegend}}; 
\end{tikzpicture}%
    \vspace{-0.5cm}
    \caption{Detection delays of QT-EWMA-update ($\beta=5$) stopping the update after acquiring $S=512,1024$ samples, compared to QT-EWMA-update, QT-EWMA over Gaussian datastreams ($d=16$) with change points at $\tau=250,500,750,1000$ with sKL$(\phi_0,\phi_1)=2$. We set initial training set sizes $N=64,128,256$ and target $\arl{0}=2000$. We observe that QT-EWMA-update outperforms QT-EWMA, and that stopping the update improves the performance by reducing the risk of updating when $t>\tau$.
    }
    \label{fig:stop_update}
\end{figure*}

On the UCI+credit datasets (Fig. \ref{fig:uci_arl} (e,f,g,h)), QT-EWMA-update is the second-best performing method (slightly outperformed by SPLL-CPM) when the training set is small, and QT-EWMA is the best-performing method when $N=4096$. On the INSECTS dataset (Fig. \ref{fig:insects_arl} (e,f,g,h)), QT-EWMA-update achieves the best detection delays when $N=64,128,256$, and QT-EWMA approaches the performance of Scan-B when the training set is large ($N=4096$). Moreover, QT-EWMA and QT-EWMA-update substantially outperform SPLL and SPLL-CPM, meaning that a QuantTree histogram can model the distribution of the INSECTS datasets much better than a Gaussian mixture. Remarkably, QT-EWMA and QT-EWMA-update consistently outperform QuantTree in all the considered scenarios, confirming that our sequential statistics are more powerful than the batch-wise Pearson statistic computed online.

As expected, the detection power of the methods based on QuantTree and Scan-B increases significantly with the training set size $N$. In contrast, SPLL-CPM achieves similar performance at different values of $N$ since the Gaussian mixture model $\widehat\phi_0$ fitted on $TR$ is sufficiently accurate even when $N$ is small, and the CPM does not require a training set \cite{Ross2011}. Most remarkably, QT-EWMA and QT-EWMA-update outperform Scan-B when $N$ is small, meaning that in these settings our online statistics have higher detection power than monitoring sliding windows by the MMD statistic. Finally, we observe that QT-EWMA-update substantially outperforms QT-EWMA when the training set is extremely small ($N=64,128$), while it yields only a marginal improvement when $N=256$, meaning that when $N\geq256$ the values $\tilde\pi_j$ are sufficiently good estimates of $p_j$.

In all the considered scenarios, QT-EWMA, QT-EWMA-update and SPLL-CPM approach the target false alarm rates computed by \eqref{eq:geosum}. In contrast, QuantTree and SPLL have, respectively, lower and higher false alarm rates than the target in all the considered monitoring scenarios. This is due to the fact that the empirical $\arl{0}$ of QuantTree is higher than the target as in Proposition \ref{prop:arl_batch}, while the empirical $\arl{0}$ of SPLL is lower than the target due to inaccurate threshold estimation, as observed previously. The false alarm rates of Scan-B, instead, exhibit a completely different behavior, which also depends on $\phi_0$ since its thresholds do not yield a constant false alarm probability.

\noindent\textbf{Stopping the update.} As discussed in Section \ref{subsec:discussion}, we can stop the update of the QuantTree histogram after acquiring a sufficient amount of data. This mitigates the problem of updating the estimated bin probabilities $\hat p_{j,t}$ when $t>\tau$, i.e., when $x_t\sim\phi_1$. To demonstrate this, we measure the detection delay of QT-EWMA-update where we stop the update after analyzing $S$ samples, i.e., when $N+t=S$. In this experiment, we compare QT-EWMA-update stopping at $S=512,1024$ against QT-EWMA and QT-EWMA-update. Fig. \ref{fig:stop_update} shows the detection delays achieved on Gaussian datastreams with $d=16$ and length $L=10000$ containing a change point at $\tau=250,500,750,1000$. We consider different training set sizes $N=64,128,256$, and set target $\arl{0}=2000$. 
As observed in the previous experiments, when $N=64,128$, QT-EWMA-update performs better than QT-EWMA, while the two algorithms have similar results when $N=256$, confirming that updating the histogram is not necessary when $N$ is sufficiently large. In all cases, the detection delay of QT-EWMA-update decreases when the change occurs later in the datastream since more samples $x_t\sim\phi_0$ are used to update $\hat p_{j,t}$. 

The fact that the detection delays of QT-EWMA-update with stopping rule are lower than those of QT-EWMA-update confirms that reducing the amount of samples $x_t\sim\phi_1$ used to update $\hat p_{j,t}$ is beneficial. The detection performance of QT-EWMA-update with stopping rule improves when the change point $\tau$ occurs later in the datastream, unless the change occurs after having stopped the update, i.e. when $\tau> S-N$. In Fig. \ref{fig:stop_update}(a,b) we observe that setting $S=512$ yields similar detection delays when $\tau=500,750,1000$ since these changes occur after having stopped the update, thus all the $S-N$ samples used to update $\hat p_{j,t}$ are drawn from $\phi_0$. In contrast, when $\tau=250$ the detection delay is higher since samples from $\phi_1$ might bias the estimates $\hat p_{j,t}$. We observe the same effect in Fig. \ref{fig:stop_update}(c) for $S=1024$, where the detection delays for $\tau=750,1000$ are very similar and lower than those obtained when $\tau=250,500$. 

\section{Conclusions}
\label{sec:conclusions}
We introduce QT-EWMA, a novel nonparametric online change-detection algorithm for multivariate datastreams. Our solution is efficient and effectively controls the $\arl{0}$ and false alarm rates, which is very useful in practical applications. We also design an updating scheme for QT-EWMA and implement QT-EWMA-update. Here we update the estimated bin probabilities of the QuantTree histogram online, as soon as new data becomes available, enabling monitoring using a very small training set. Our experiments on synthetic and real-world datastreams show that alternative solutions do not provide such guarantees in nonparametric settings, and that QT-EWMA and QT-EWMA-update achieve excellent performance, especially on real-world data.  

\bibliographystyle{IEEEtran}
\bibliography{biblio}


\begin{IEEEbiography}[{\includegraphics[width=1in,height=1.25in,clip,keepaspectratio]{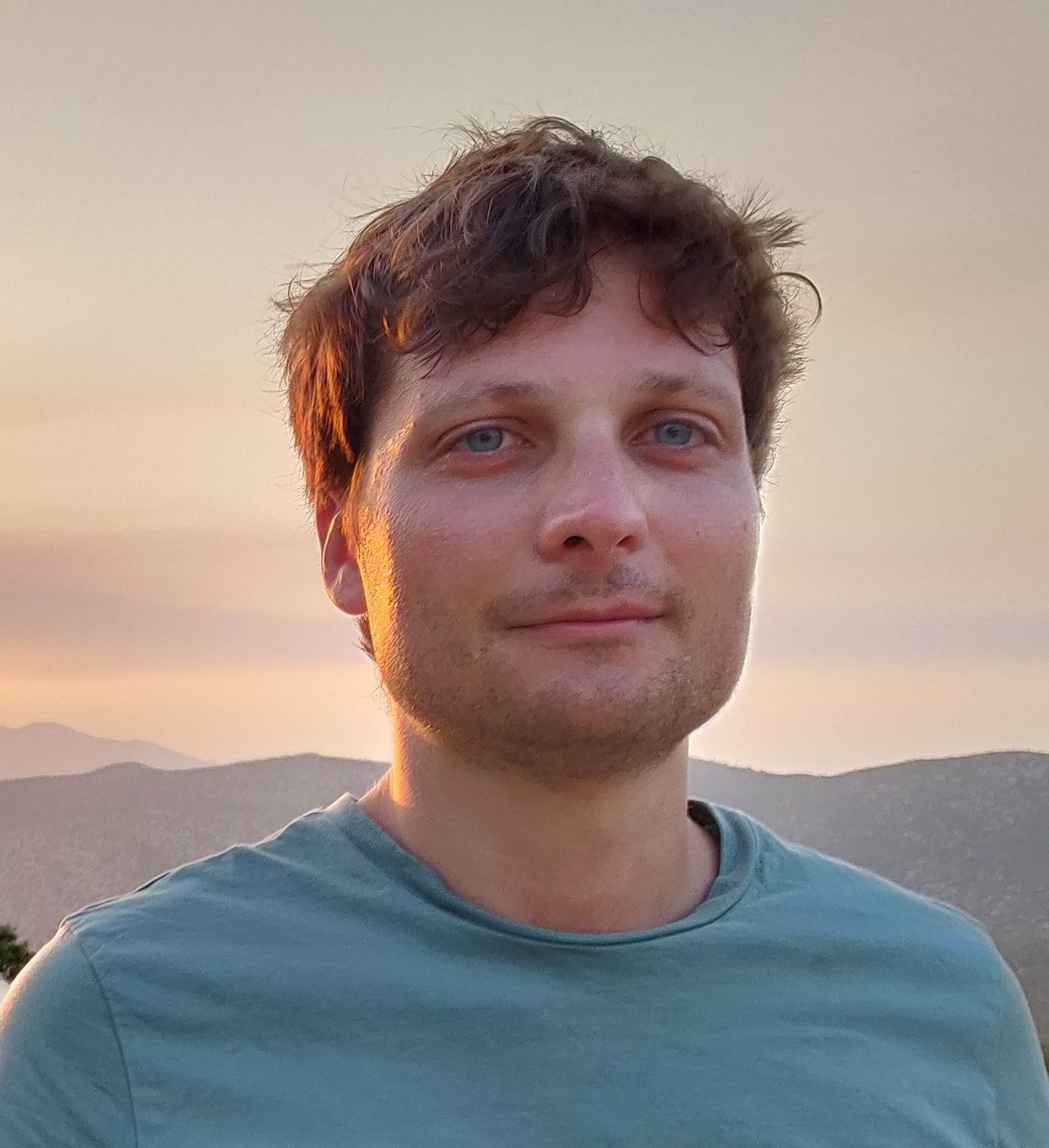}}]{Luca Frittoli} graduated in Mathematics at  Università degli Studi di Milano in 2018, and is currently working towards the Ph.D. in Information Technology at Politecnico di Milano. His research interests include change detection in multivariate datastreams, concept-drift detection, and deep learning methods for anomaly detection and open-set recognition in images and point clouds.
\end{IEEEbiography}

\begin{IEEEbiography}[{\includegraphics[width=1in,height=1.25in,clip,keepaspectratio]{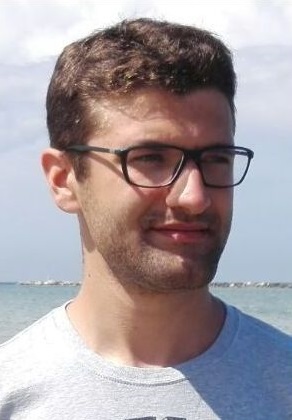}}]{Diego Carrera}
Diego Carrera graduated in Mathematics at Università degli Studi di Milano in 2013 and received the Ph.D. in Information Technology in 2018. In 2015 he has been visiting researcher at the Tampere University of Technology. Currently he is an Application Development Engineer at STMicroelectronics, where he is developing quality inspection systems to monitor the wafer production. His research interests are mainly focused on unsupervised learning algorithms, in particular change detection in high dimensional datastreams, anomaly detection in signal and images, and domain adaptation.
\end{IEEEbiography}
\begin{IEEEbiography}[{\includegraphics[width=1in,height=1.25in,clip,keepaspectratio]{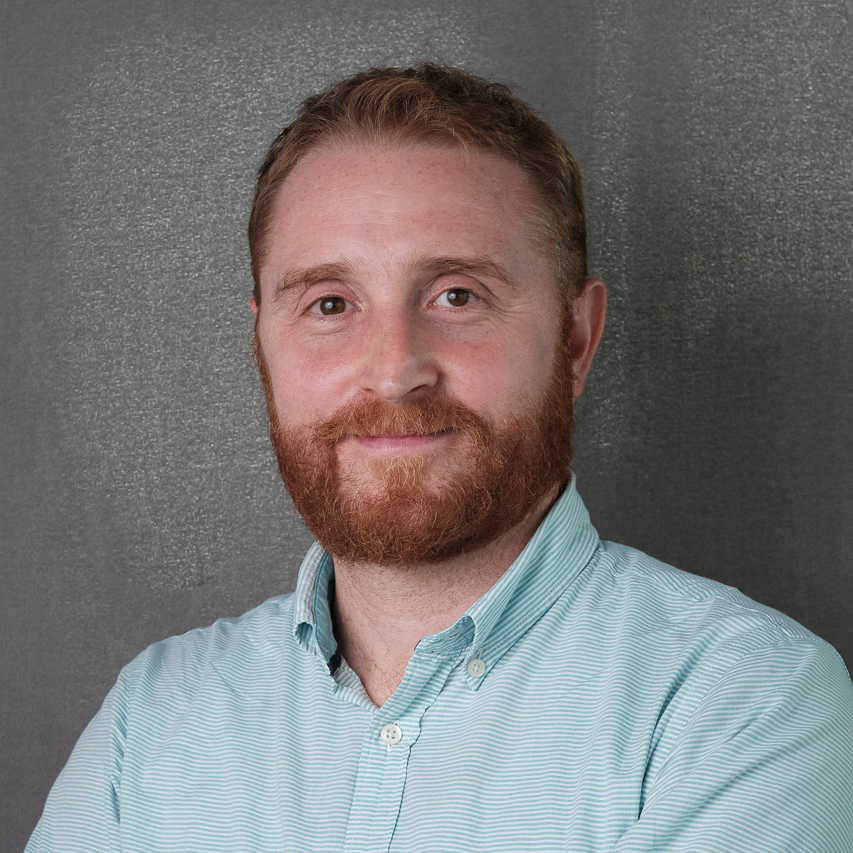}}]{Giacomo Boracchi} is an Associate Professor of Computer Engineering at Politecnico di Milano - DEIB, where he also received the Ph.D. in Information Technology (2008), after graduating in Mathematics (Università degli Studi di Milano, 2004).  His research interests concern machine learning and image processing, and in particular change/anomaly detection, domain adaptation, image restoration and analysis. Since 2015 he is leading industrial research projects concerning outlier detection systems, X-ray systems, and automatic quality inspection systems.
He has published more than 70 papers in international conferences and journals, he is currently associate editor for IEEE Transactions on Image Processing and in 2019 - 2020 he served as an associate editor for IEEE Computational Intelligence Magazine. In 2015 he received an IBM Faculty Award, in 2016 the IEEE Transactions on Neural Networks and Learning Systems Outstanding Paper Award, in 2017 the Nokia Visiting Professor Scholarship, and in 2021 the NVIDIA Applied Research Grant.  He has held tutorials in major IEEE conferences: ICIP 2020, ICASSP 2018 and IJCNN 2017 and 2019.
\end{IEEEbiography}

\end{document}


\definecolor{color_qt}{rgb}{.5,.8,.3}
\definecolor{color_l}{rgb}{.2,.6,.8}
\definecolor{color_h}{rgb}{1,.6,.1}

\definecolor{our_blue}{rgb}{.2,.6,.8}
\definecolor{our_orange}{rgb}{1,.5,.05}
\definecolor{our_green}{rgb}{.17,.63,.17}
\definecolor{our_red}{rgb}{.84,.15,.16}
\definecolor{our_purple}{rgb}{.58,.4,.74}
\definecolor{our_msblue}{rgb}{.48,.41,.93}
\definecolor{our_mblue}{rgb}{0,0,1}
\definecolor{our_darkblue}{rgb}{0,0,.55}
\definecolor{our_indigo}{rgb}{.29,0,.51}

\pgfplotsset{
    style_stop/.style={
        width=0.24\textwidth,
        height=0.26\columnwidth,
        scale only axis,
        xtick = {250,500,750,1000}, 
        title style = {
            font = \scriptsize,
            at = {(0.5, 0.95)},
        },
        xlabel style = {
        	align=center,
        	anchor = near ticklabel, 
        	at = {(0.5,-0.1)},
            font=\scriptsize,
        },
        xticklabel style = {font = \tiny},
        yticklabel style = {font = \tiny},
        ylabel style = {
            anchor = center, 
            at = {(-0.18,0.5)},
            font=\scriptsize,
        },
        /pgf/number format/.cd,
        1000 sep={},
    }
}

\pgfplotsset{
    style_auc/.style={
        width=0.24\textwidth,
        height=0.26\columnwidth,
        scale only axis,
        xtick = {1,2,3,4,5},
        xticklabels = {$50$,$100$,$200$,$500$,$1000$}, 
        title style = {
            font = \scriptsize,
            at = {(0.5, 0.95)},
        },
        xlabel style = {
        	align=center,
        	anchor = near ticklabel, 
        	at = {(0.5,-0.1)},
            font=\scriptsize,
        },
        xticklabel style = {font = \tiny},
        yticklabel style = {font = \tiny},
        ylabel style = {
            anchor = center, 
            at = {(-0.18,0.5)},
            font=\scriptsize,
        },
        ymajorgrids=true,
    }
}

\pgfplotsset{
    style_skl/.style={
        width=0.24\textwidth,
        height=0.13\textwidth,
        scale only axis,
        xtick = {0.5,1,...,3}, 
        xticklabels = {$0.5$,$1$,$1.5$,$2$,$2.5$,$3$},
        title style = {
            font = \scriptsize,
            at = {(0.5, 0.95)},
        },
        xlabel style = {
        	align=center,
        	anchor = near ticklabel, 
        	at = {(0.5,-0.1)},
            font=\scriptsize,
        },
        xticklabel style = {font = \tiny},
        yticklabel style = {font = \tiny},
        ylabel style = {
            anchor = center, 
            at = {(-0.18,0.5)},
            font=\scriptsize,
        },
    }
}

\pgfplotsset{
    style_bins/.style={
        width=0.38\textwidth,
        height=0.33\columnwidth,
        scale only axis,
        xtick = {0,1,...,8},
        xticklabels = {$2$,$4$,$8$,$16$,$32$,$64$,$128$,$256$,$512$},
        title style = {
            font = \scriptsize,
            at = {(0.5, 0.95)},
        },
        xlabel style = {
        	align=center,
        	anchor = near ticklabel, 
        	at = {(0.5,-0.1)},
            font=\scriptsize,
        },
        xticklabel style = {font = \tiny},
        yticklabel style = {font = \tiny},
        ylabel style = {
            anchor = center, 
            at = {(-0.12,0.5)},
            font=\scriptsize,
        },
    }
}

\pgfplotsset{
    style_arl/.style={
        width=0.19\textwidth,
        height=0.135\textwidth,
        scale only axis,
        scaled ticks=false,
        xmin = 175, 
        xmax = 5375,
        xtick = {500,1000,2000,5000},
        xticklabels = {$500\;\;$,$\;\;1000$,$\;2000$,$5000$},
        title style = {
            font = \scriptsize,
            at = {(0.5, 0.95)},
        },
        xlabel style = {
        	align=center,
        	anchor = near ticklabel, 
        	at = {(0.5,-0.1)},
            font=\scriptsize,
        },
        xticklabel style = {font = \tiny},
        yticklabel style = {font = \tiny},
        ylabel style = {
            anchor = center, 
            at = {(-0.22,0.5)},
            font=\scriptsize,
        },
        /pgf/number format/.cd,
        use comma,
        1000 sep={},
    }
}

\pgfplotsset{
    style_delay/.style={
        width=0.19\textwidth,
        height=0.135\textwidth,
        scale only axis,
        xtick = {0,10,...,80}, 
        minor xtick = {63, 39, 22, 9.5},
        title style = {
            font = \scriptsize,
            at = {(0.5, 0.95)},
        },
        xlabel style = {
        	align=center,
        	anchor = near ticklabel, 
        	at = {(0.5,-0.1)},
            font=\scriptsize,
        },
        xticklabel style = {font = \tiny},
        yticklabel style = {font = \tiny},
        ylabel style = {
            anchor = center, 
            at = {(-0.22,0.5)},
            font=\scriptsize,
        },
        /pgf/number format/.cd,
        use comma,
        1000 sep={},
    }
}

\pgfplotsset{
    style_fig_del_sim/.style={
        width=0.17\textwidth,
        height=0.23\columnwidth,
        scale only axis,
        xmin = 0.2, 
        xmax = 3.3,
        xtick = {0.5,1,...,3}, 
        title style = {
            font = \scriptsize,
            at = {(0.5, 0.9)},
        },
        xlabel style = {
        	align=center,
        	anchor = near ticklabel, 
        	at = {(0.5,-0.15)},
            font=\footnotesize,
        },
        xticklabel style = {font = \scriptsize},
        yticklabel style = {font = \scriptsize},
        ylabel style = {
            anchor = center, 
            at = {(-0.25,0.5)},
            font=\footnotesize,
        },
    }
}

\pgfplotsset{
    style_fig_del_real/.style={
        width=0.175\textwidth,
        height=0.31\columnwidth,
        scale only axis,
        xmin = 0, 
        xmax = 1,
        xtick = {0.001,0.01,0.1,1}, 
        title style = {
            font = \small,
            at = {(0.5, 0.9)},
        },
        xlabel style = {
        	align=center,
        	anchor = near ticklabel, 
        	at = {(0.5,-0.1)},
            font=\footnotesize,
        },
        xticklabel style = {font = \scriptsize},
        yticklabel style = {font = \scriptsize},
        ylabel style = {
            anchor = center, 
            at = {(-0.25,0.5)},
            font=\footnotesize,
        },
    }
}

\pgfplotsset{
    style_line/.style={
        line width = 1pt,
    },
}

\pgfplotsset{
    style_qtewma/.style={
        style_line,
        color = our_blue,
        mark = square,
    },
    style_qt/.style={
        style_line,
        color = our_orange,
        mark = o,
    },
    style_spll/.style={
        style_line,
        color = our_red,
        mark = asterisk,
    },
    style_cpm/.style={
        style_line,
        color = our_green,
        mark = diamond,
    },
    style_scanb/.style={
        style_line,
        color = our_purple,
        mark = triangle,
    },
    style_truep/.style={
        dashed,
        style_line,
        color = our_orange,
        mark = star,
    },
    style_beta2/.style={
        style_line,
        color = our_green,
        mark = halfsquare*,
    },
    style_beta5/.style={
        style_line,
        color = our_mblue,
        mark = pentagon,
    },
    style_beta10/.style={
        style_line,
        color = our_red,
        mark = halfcircle*,
    },
    style_stop512/.style={
        style_line,
        color = our_green,
        mark = x,
    },
    style_stop1024/.style={
        style_line,
        color = our_red,
        mark = triangle*,
    }
}

\IEEEraisesectionheading{
\section*{Supplementary Material}
}

Here we provide the proofs omitted in the main article due to space limitations (Section \ref{sec:proof}) and additional figures and comments that confirm our experimental results (Section \ref{sec:more_experiments}).

\section{Proofs}\label{sec:proof}
\begin{proposition}\label{prop:arl_batch}
Let $W_t$ be any batch of $\nu$ samples drawn from $\phi_0$ and let the detection threshold $h^{\nu}$ be such that 
\begin{equation}\label{eq:fpr}
\mathbb P_{\phi_0}(T^{\nu}(W_t) > h^{\nu}) = \alpha.
\end{equation}
Then, the monitoring scheme $T^{\nu}(W_t) > h^{\nu}$ yields $\arl{0}\geq \nu/\alpha$.
\end{proposition}
\begin{proof}
Let $t_B^*$ the first time instant such that $T^{\nu}(W_{t_B^*}) > h$ and let us compute $\arl{0}=\mathbb E[t_B^*]$. To this purpose, we follow a strategy similar to that in Section 4.2. At first we observe that since the batches does not overlap, the variables $\{T^{\nu}(W_t)\}$ are independent if we condition w.r.t. the specific training set realization (thus the model used to compute $T^{\nu}$). Therefore, we obtain that:
\begin{equation}\label{eq:cond_indipendence}
    \begin{aligned}
    \mathbb P(T^{\nu}(W_t)>h \;|\; TR, T^{\nu}(W_k)\leq h\; \forall k < t) = \\ = \mathbb P(T^{\nu}(W_t)>h \;|\; TR).
    \end{aligned}
\end{equation}
Let us define the random variable $p=\mathbb P(T^{\nu}(W)>h \;|\; TR)$, where $W$ is a batch of $\nu$ samples drawn from $\phi_0$. Following \cite{margavio1995alarm}, the random variable $t_B^*$ is distributed as a geometric random variable w.r.t. the conditional probability $\mathbb P(\cdot | TR)$, and its expected value is 
\begin{equation}
    \mathbb E[t_B^* \;|\; TR] = \frac{1}{p}.
\end{equation}
To compute the $\arl{0}$ we only have to evaluate the expectation of $1/p$ w.r.t. to the training set realizations since the law of total expectation implies that
\begin{equation}
    \mathbb E[t_B^*] = \mathbb E[\mathbb E[t_B^* \;|\; TR]] = \mathbb E\left[\frac{1}{p}\right].
\end{equation}
We observe that Jensen's inequality implies that
\begin{equation}\label{eq:jensen}
    \mathbb E[t_B^*] = \mathbb E\left[\frac{1}{p}\right] \geq \frac{1}{\mathbb E[p]}, 
\end{equation}
since the function $1/p$ is convex for $p>0$. Finally, we have to compute $\mathbb E[p]$. To this purpose we rewrite
\begin{equation}
    p = \mathbb P(T^{\nu}(W)>h^{\nu} \;|\; TR) = \mathbb E[\mathds{1}(T^{\nu}(W)>h^{\nu}) \;|\; TR],
\end{equation}    
where $\mathds 1$ denotes the indicator function. Then:
\begin{align}
    \label{eq:exp_p}\mathbb E[p] &= \mathbb E[\mathbb E[\mathds{1}(T^{\nu}(W)>h^{\nu}) | TR]] =\\
    \label{eq:tower}&=\mathbb E[\mathds{1}(\{T^{\nu}(W)>h^{\nu}\})] = \\ &= \mathbb P(T^{\nu}(W_t)>h^{\nu}) = \alpha,
\end{align}
where the equality between \eqref{eq:exp_p} and \eqref{eq:tower} is due to the law of total expectation. Combining \eqref{eq:exp_p} and \eqref{eq:jensen} we obtain that
\begin{equation}
    \mathbb E[t_B^*] \geq \frac{1}{\alpha}.
\end{equation}
To obtain the thesis we observe that, since the monitoring is performed in a batch-wise manner, change detected after the $t_B^*$ batch translates in a detection made after $\nu\cdot t_B^*$ samples of the datastream, so $\arl{0}\geq \nu/\alpha$.
\end{proof}

\begin{proposition}\label{prop:exact}
Let $W_t$ be any batch of $\nu$ samples drawn from $\phi_0$ and let the detection threshold $h^{\nu}$ be such that 
\begin{equation}\label{eq:fpr_bootstrap}
\mathbb P_{\phi_0}(T^{\nu}(W_t) > h^{\nu} \;|\; TR) = \alpha,
\end{equation}
Then, the monitoring scheme $T^{\nu}(W_t) > h^{\nu}$ yields $\arl{0}= \nu/\alpha$.
\end{proposition}

\begin{proof}
Following the same strategy we pursued to prove Proposition \ref{prop:arl_batch}, we have that the random variable $p = \mathbb P(T^{\nu}(W_t)>h^{\nu} | TR)$ is a constant equal to $\alpha$. Therefore, the equality holds in \eqref{eq:jensen}, from which we derive $\arl{0}= \nu/\alpha$.
\end{proof}

\begin{figure*}[t!]
    \centering
    \input{new_plots/main_4}
    \caption{Experimental results over Gaussian datastreams ($d=4$), using training sets of size $N=64,128,256,4096$. (a,b,c,d) show the empirical $\arl{0}$ of the considered methods compared to the target. We observe that the empirical $\arl{0}$ of QT-EWMA, QT-EWMA-update and SPLL-CPM approaches the target, while Scan-B and SPLL cannot maintain the target $\arl{0}$. (e,f,g,h) show the detection delay plotted against the false alarm rate with change point $\tau=500$ and sKL$(\phi_0,\phi_1)=1$. In terms of detection delay, the best methods are QT-EWMA-update and SPLL-CPM when using small training sets ($N=64,128,256$) and SPLL and Scan-B when using large training sets ($N=4096$). We observe that QT-EWMA-update clearly outperforms QT-EWMA when $N=64,128$, and that only QT-EWMA, QT-EWMA-update and SPLL-CPM achieve the target false alarm rates given by \eqref{eq:geosum}, which are represented in the plots by vertical dotted lines.}
    \label{fig:gaussian4}
\end{figure*}

\begin{figure*}[t!]
    \centering
    \begin{tikzpicture}
\begin{groupplot}[ 
    group style={group size=3 by 1,}, 
    width=0.28\textwidth,
    height=0.135\textwidth,
    title style = {font=\scriptsize,
    },
    scale only axis,
    ymajorgrids,
    xlabel style = {anchor = near ticklabel, 
        	at = {(0.5,-0.1)},font=\scriptsize},
    ylabel style = {font=\scriptsize},
    xticklabel style = {font = \tiny},
    yticklabel style = {font = \tiny},
    xtick = {0.5,1,...,3},
]
%
\nextgroupplot[
        title={Gaussian, $d=4,N=4096$},
        ylabel={detection delay},
        xlabel={sKL},
        ytick = {0,100,...,600},
        /pgf/number format/.cd,use comma,1000 sep={},scale only axis,scaled ticks=false,]
\addplot [style_qtewma,]
table [
    x=skl,
    y=qt_ewma,
]{data/delay_skl_4.txt};

\addplot [style_qt,]
table [
    x=skl,
    y=qt_batch,
]{data/delay_skl_4.txt};

\addplot [style_spll,]
table [
    x=skl,
    y=spll,
]{data/delay_skl_4.txt};

\addplot [style_cpm,]
table [
    x=skl,
    y=spll_cpm,
]{data/delay_skl_4.txt};

\addplot [style_scanb,]
table [
    x=skl,
    y=Scan-B,
]{data/delay_skl_4.txt};

\nextgroupplot[
        title={Gaussian, $d=16,N=4096$},
        xlabel={sKL},
        ytick = {0,200,...,1000},
        /pgf/number format/.cd,use comma,1000 sep={},scale only axis,scaled ticks=false,]
\addplot [style_qtewma,]
table [
    x=skl,
    y=qt_ewma,
]{data/delay_skl_16.txt};

\addplot [style_qt,]
table [
    x=skl,
    y=qt_batch,
]{data/delay_skl_16.txt};

\addplot [style_spll,]
table [
    x=skl,
    y=spll,
]{data/delay_skl_16.txt};

\addplot [style_cpm,]
table [
    x=skl,
    y=spll_cpm,
]{data/delay_skl_16.txt};

\addplot [style_scanb,]
table [
    x=skl,
    y=Scan-B,
]{data/delay_skl_16.txt};

\nextgroupplot[
        title={Gaussian, $d=64,N=4096$},
        xlabel={sKL},
        ytick = {0,250,...,1500},
        legend style={at={($(0,0)+(0cm,0cm)$)},legend columns=-1,fill=none,draw=black,anchor=center,align=center, font=\scriptsize, legend to name = grouplegend,},
        /pgf/number format/.cd,use comma,1000 sep={},scale only axis,scaled ticks=false,]
\addplot [style_qtewma,]
table [
    x=skl,
    y=qt_ewma,
]{data/delay_skl_64.txt};
\addlegendentry{QT-EWMA}

\addplot [style_qt,]
table [
    x=skl,
    y=qt_batch,
]{data/delay_skl_64.txt};
\addlegendentry{QuantTree~\cite{boracchi2018quanttree}}

\addplot [style_spll,]
table [
    x=skl,
    y=spll,
]{data/delay_skl_64.txt};
\addlegendentry{SPLL~\cite{kuncheva2011change}}

\addplot [style_cpm,]
table [
    x=skl,
    y=spll_cpm,
]{data/delay_skl_64.txt};
\addlegendentry{SPLL-CPM}

\addplot [style_scanb,]
table [
    x=skl,
    y=Scan-B,
]{data/delay_skl_64.txt};
\addlegendentry{Scan-B~\cite{li2015m}}

\end{groupplot}
\node[align=center] at ($(group c1r1.south west)!.5!(group c1r1.south east) + (0cm,-0.8cm)$) {\scriptsize(a)};
\node[align=center] at ($(group c2r1.south west)!.5!(group c2r1.south east) + (0cm,-0.8cm)$) {\scriptsize(b)};
\node[align=center] at ($(group c3r1.south west)!.5!(group c3r1.south east) + (0cm,-0.8cm)$) {\scriptsize(c)};
\node at ($(group c2r1.south west)!.5!(group c2r1.south east) + (0cm,-1.5cm)$) {\ref{grouplegend}}; 
\end{tikzpicture}%
    \vspace{-0.5cm}
    \caption{Detection delays of the considered methods over Gaussian datastreams ($d=4,16,64$) with different change magnitudes $\text{sKL}(\phi_0,\phi_1)=0.5,1,2.5,2,2.5,3$. For a fair comparison, we set target $\arl{0}=1000$, which is maintained by all methods (see Fig. \ref{fig:gaussian4}(d)). As expected, all methods improve their performance when increasing $\text{sKL}(\phi_0,\phi_1)$ and achieve higher detection delays increasing $d$. We observe that when $d=4$ QT-EWMA is on par with SPLL, SPLL-CPM and Scan-B, which outperform QT-EWMA when $d=16,64$.}
    \label{fig:skl}
\end{figure*}

\section{Additional Experiments}\label{sec:more_experiments}


\noindent\textbf{Empirical $\arl{0}$.} The comparison between the empirical and target $\arl{0}$ on simulated Gaussian datastreams ($d=4$) is reported in Fig. \ref{fig:gaussian4} (a,b,c,d), which confirms that QT-EWMA, QT-EWMA-update and SPLL-CPM control the $\arl{0}$ very accurately, independently from the data dimension $d$ and the training set size $N$, as shown in the main article. The empirical $\arl{0}$ of QT is higher than the target, as we expected from Proposition \ref{prop:arl_batch}, while Scan-B cannot maintain high target $\arl{0}$. We obtain the same results on datastreams sampled from the credit \cite{dal2017credit} (see Fig. \ref{fig:credit} (a,b,c,d)) and all the considered UCI datasets \cite{uci2019} (see Fig. \ref{fig:sensorless}-\ref{fig:lodgepole}), confirming the nonparametric nature of QuantTree and the limitations of Scan-B in maintaining the target $\arl{0}$.

\noindent\textbf{Detection delay vs false alarms.} Fig. \ref{fig:gaussian4} (e,f,g,h) shows the detection delays of the considered methods on simulated Gaussian datastreams ($d=4$) containing a change point at $\tau=500$, plotted against their false alarm rates. In this experiment we set target $\arl{0}=500,1000,2000,5000$. In terms of detection delay, when the training set is small ($N=64,128,256$) QT-EWMA-update is on par with SPLL-CPM, which represents the best-performing method in our experiments over higher-dimensional Gaussian datastreams ($d=16,64$), as shown in the main article. When the training set is large ($N=4096$) SPLL and Scan-B outperform QT-EWMA on Gaussian datastreams, confirming the results presented in the main article. 

Since on Gaussian data we can control the change magnitude by setting sKL$(\phi_0,\phi_1)$ \cite{carrera2018generating}, in Fig. \ref{fig:skl} we show the detection delays of the considered methods depending on the change magnitude $\text{sKL}(\phi_0,\phi_1)=0.5,1,1.5,2,2.5,3$. In this experiment we use large training sets ($N=4096$) and set target $\arl{0}=1000$, which is maintained by all methods (see Fig. \ref{fig:gaussian4}(d)), to make the comparison fair. As expected, the detection delays of all methods decrease when the change magnitude increases, i.e., when the distribution change becomes more evident. All methods achieve higher detection delays as $d$ increases, which is also expected due to \emph{detectability loss} \cite{alippi2015change}. In particular, in Fig. \ref{fig:skl} (a) we observe that when $d=4$ QT-EWMA is on par with the best-performing methods, i.e., the parametric SPLL and SPLL-CPM and the nonparametric Scan-B. In contrast, in Fig. \ref{fig:skl} (b,c) we can see that SPLL and Scan-B have lower detection delays than QT-EWMA when $d=16,64$, confirming that statistics based on verified parametric assumptions (SPLL) and on Maximum Mean Discrepancy (Scan-B) are very powerful and yield change-detection algorithms that are more robust to detectability loss, as observed in the main article.

When the training set is small ($N=64,128,256$) QT-EWMA-update is the best method in terms of detection delay over the particle (Fig. \ref{fig:particle} (e,f,g)) and ni\~no (Fig. \ref{fig:nino} (e,f,g)) datasets, and is outperformed only by SPLL-CPM over the other datasets, though only slightly over spruce (Fig. \ref{fig:spruce} (e,f,g)) and lodgepole (Fig. \ref{fig:lodgepole} (e,f,g)). This confirms that, overall, SPLL-CPM slightly outperforms QT-EWMA-update over the UCI+credit datasets when $N=64,128,256$, as shown by the average results presented in the main article. In contrast, Scan-B yields higher detection delays compared to QT-EWMA-update on all the considered datasets when $N=64,128,256$.

\begin{figure*}[t!]
    \centering
    \input{new_plots/main_credit}
    \vspace{-0.5cm}
    \caption{Experimental results over the credit dataset ($d=28$) \cite{dal2017credit}, using training sets of size $N=64,128,256,4096$. (a,b,c,d) show the empirical $\arl{0}$ of the considered methods compared to the target. We observe that the empirical $\arl{0}$ of QT-EWMA, QT-EWMA-update and SPLL-CPM approaches the target, while Scan-B and SPLL cannot maintain the target $\arl{0}$. (e,f,g,h) show the detection delay plotted against the false alarm rate with change point $\tau=500$. In terms of detection delay, the best method is SPLL-CPM when $N=64,128$, QT-EWMA and QT-EWMA-update outperform all the other methods when $N=256$, while Scan-B is the best method when $N=4096$. We observe that only QT-EWMA, QT-EWMA-update and SPLL-CPM achieve the target false alarm rates given by \eqref{eq:geosum}, which are represented in the plots by vertical dotted lines.}
    \label{fig:credit}
\end{figure*}

\begin{figure*}[t!]
    \centering
    \input{new_plots/main_sensorless}
    \vspace{-0.5cm}
    \caption{Experimental results over the sensorless dataset ($d=48$) \cite{uci2019}, using training sets of size $N=64,128,256,4096$. (a,b,c,d) show the empirical $\arl{0}$ of the considered methods compared to the target. We observe that the empirical $\arl{0}$ of QT-EWMA, QT-EWMA-update and SPLL-CPM approaches the target, while Scan-B and SPLL cannot maintain the target $\arl{0}$. (e,f,g,h) show the detection delay plotted against the false alarm rate with change point $\tau=500$. In terms of detection delay, SPLL-CPM is the best method in all the considered scenarios, and SPLL achieves similar results when $N=4096$. We observe that only QT-EWMA, QT-EWMA-update and SPLL-CPM achieve the target false alarm rates given by \eqref{eq:geosum}, which are represented in the plots by vertical dotted lines.}
    \label{fig:sensorless}
\end{figure*}

As observed in the main article, QT-EWMA-update outperforms QT-EWMA in terms of detection delay when the training set is small (especially when $N=64,128$) over both synthetic (Fig. \ref{fig:gaussian4} (e,f,g)) and real-world datastreams (Fig. \ref{fig:credit}-\ref{fig:lodgepole}). These results confirm that updating the QuantTree histograms improves the detection performance when the initial training set is small. Moreover, over all the considered datasets we have that QT-EWMA and QT-EWMA-update outperform QT, consistently with the experiments illustrated in the main article.

\begin{figure*}[t!]
    \centering
    \input{new_plots/main_particle}
    \vspace{-0.5cm}
    \caption{Experimental results over the particle dataset ($d=50$) \cite{uci2019}, using training sets of size $N=64,128,256,4096$. (a,b,c,d) show the empirical $\arl{0}$ of the considered methods compared to the target. We observe that the empirical $\arl{0}$ of QT-EWMA, QT-EWMA-update and SPLL-CPM approaches the target, while Scan-B and SPLL cannot maintain the target $\arl{0}$. (e,f,g,h) show the detection delay plotted against the false alarm rate with change point $\tau=500$. In terms of detection delay, the best method is QT-EWMA-update when $N=64,128,256$, and QT-EWMA when $N=4096$. We observe that only QT-EWMA, QT-EWMA-update and SPLL-CPM achieve the target false alarm rates given by \eqref{eq:geosum}, which are represented in the plots by vertical dotted lines.}
    \label{fig:particle}
\end{figure*}

When the training set is large ($N=4096$), Scan-B substantially improves its detection delays, being the best-performing method over the credit dataset (Fig. \ref{fig:credit} (h)) and approaching QT-EWMA over the sensorless (Fig. \ref{fig:sensorless} (h)) and particle (Fig. \ref{fig:particle} (h)) datasets. QT-EWMA outperforms all the other methods over the particle (Fig. \ref{fig:particle} (h)) and ni\~no (Fig. \ref{fig:nino} (h)) datasets, while SPLL-CPM yields the lowest detection delays on the remaining datasets, although it only slightly outperforms QT-EWMA over the protein (Fig. \ref{fig:protein} (h)), spruce (Fig. \ref{fig:spruce} (h)) and lodgepole (Fig. \ref{fig:lodgepole} (h)) datasets. These plots show that QT-EWMA is very effective on the UCI+credit datasets in terms of detection delay, confirming the average results reported in the main article. Moreover, these experimental results confirm our observation that Scan-B requires a large training set to achieve good detection performance.

We recall that, when the false alarm probability of a change-detection algorithm is a constant $\alpha$ at each time $t$, its stopping time $t^*$ is a Geometric random variable having parameter $\alpha$ \cite{margavio1995alarm}. Hence, when the $\arl{0}$ is controlled by setting a constant false alarm probability $\alpha=1/\arl{0}$, the false alarm probability at any time $t$ corresponds to the Geometric sum:
\begin{equation}\label{eq:geosum}
    \mathbb{P}_{\phi_0}(t^*\leq t) = \sum_{k=1}^t \alpha(1-\alpha)^{k-1} = 1 - (1-\alpha)^t,
\end{equation}
thus the change-detection algorithm can control the false alarm rate at any time $t$, as we show in the main article. The results presented here over Gaussian datastreams with $d=4$ (Fig. \ref{fig:gaussian4} (e,f,g,h)) and over datastreams sampled from the considered UCI+credit datasets (Fig. \ref{fig:credit}-\ref{fig:lodgepole}) confirm that QT-EWMA, QT-EWMA-update and SPLL-CPM approach the target values computed by \eqref{eq:geosum} at $\arl{0}=500,1000,2000,5000$. As in the main article, we observe that QT has fewer false alarms than expected, and this is a consequence the fact that its empirical $\arl{0}$ is greater than the target due to Proposition \ref{prop:arl_batch}. In contrast, SPLL has more false alarms than expected since its empirical $\arl{0}$ is slightly lower than the target, which confirms that computing the thresholds by bootstrap over a limited training set yields inaccurate estimates, as we observe also in the main article. The false alarms of Scan-B, instead, exhibit a completely different behavior, which also depends on the data distribution, confirming that its thresholds do not yield a constant false alarm probability. All in all, these results confirm those presented in the main article, showing that QT-EWMA and QT-EWMA-update can control the false alarm rates in all the considered datasets and monitoring scenarios.

\begin{figure*}[t!]
    \centering
    \input{new_plots/main_protein}
    \vspace{-0.5cm}
    \caption{Experimental results over the protein dataset ($d=9$) \cite{uci2019}, using training sets of size $N=64,128,256,4096$. (a,b,c,d) show the empirical $\arl{0}$ of the considered methods compared to the target. We observe that the empirical $\arl{0}$ of QT-EWMA, QT-EWMA-update and SPLL-CPM approaches the target, while Scan-B and SPLL cannot maintain the target $\arl{0}$. (e,f,g,h) show the detection delay plotted against the false alarm rate with change point $\tau=500$. In terms of detection delay, the best method is SPLL-CPM for all the considered values of $N$, approached by QT-EWMA when $N=4096$. We observe that only QT-EWMA, QT-EWMA-update and SPLL-CPM achieve the target false alarm rates given by \eqref{eq:geosum}, which are represented in the plots by vertical dotted lines.}
    \label{fig:protein}
\end{figure*}

\begin{figure*}[t!]
    \centering
    \input{new_plots/main_nino}
    \vspace{-0.5cm}
    \caption{Experimental results over the ni\~no dataset ($d=5$) \cite{uci2019}, using training sets of size $N=64,128,256,4096$. (a,b,c,d) show the empirical $\arl{0}$ of the considered methods compared to the target. We observe that the empirical $\arl{0}$ of QT-EWMA, QT-EWMA-update and SPLL-CPM approaches the target, while Scan-B and SPLL cannot maintain the target $\arl{0}$. (e,f,g,h) show the detection delay plotted against the false alarm rate with change point $\tau=500$. In terms of detection delay, the best method is QT-EWMA-update when $N=64,128,256$, and QT-EWMA when $N=4096$. We observe that only QT-EWMA, QT-EWMA-update and SPLL-CPM achieve the target false alarm rates given by \eqref{eq:geosum}, which are represented in the plots by vertical dotted lines.}
    \label{fig:nino}
\end{figure*}

\begin{figure*}[t!]
    \centering
    \input{new_plots/main_spruce}
    \vspace{-0.7cm}
    \caption{Experimental results over the spruce dataset ($d=10$) \cite{uci2019}, using training sets of size $N=64,128,256,4096$. (a,b,c,d) show the empirical $\arl{0}$ of the considered methods compared to the target. We observe that the empirical $\arl{0}$ of QT-EWMA, QT-EWMA-update and SPLL-CPM approaches the target, while Scan-B and SPLL cannot maintain the target $\arl{0}$. (e,f,g,h) show the detection delay plotted against the false alarm rate with change point $\tau=500$. In terms of detection delay, the best method is SPLL-CPM, although all the methods except Scan-B perform similarly. We observe that only QT-EWMA, QT-EWMA-update and SPLL-CPM achieve the target false alarm rates given by \eqref{eq:geosum}, which are represented in the plots by vertical dotted lines.}
    \label{fig:spruce}
\end{figure*}

\begin{figure*}[t!]
    \centering
    \input{new_plots/main_lodgepole}
    \vspace{-0.7cm}
    \caption{Experimental results over the lodgepole dataset ($d=10$) \cite{uci2019}, using training sets of size $N=64,128,256,4096$. (a,b,c,d) show the empirical $\arl{0}$ of the considered methods compared to the target. We observe that the empirical $\arl{0}$ of QT-EWMA, QT-EWMA-update and SPLL-CPM approaches the target, while Scan-B and SPLL cannot maintain the target $\arl{0}$. (e,f,g,h) show the detection delay plotted against the false alarm rate with change point $\tau=500$. In terms of detection delay, the best method is SPLL-CPM, although all the methods except Scan-B perform similarly. We observe that only QT-EWMA, QT-EWMA-update and SPLL-CPM achieve the target false alarm rates given by \eqref{eq:geosum}, which are represented in the plots by vertical dotted lines.}
    \label{fig:lodgepole}
\end{figure*}

\bibliographystyle{IEEEtran}
\bibliography{biblio}